\documentclass[lettersize,journal]{IEEEtran}
\usepackage{array}
\usepackage{stfloats}
\usepackage{url}
\usepackage{verbatim}
\usepackage{cite}
\usepackage{graphicx}%
\usepackage{multirow}%
\usepackage{amsmath,amssymb,amsfonts}%
\usepackage{amsthm}%
\usepackage{mathrsfs}%
\usepackage{textcomp}%
\usepackage{manyfoot}%
\usepackage{booktabs}%
\usepackage{algorithm}%
\usepackage{algorithmicx}%
\usepackage{algpseudocode}%
\usepackage{listings}%
\usepackage{tcolorbox}
\usepackage{tabularx}
\usepackage{booktabs}
\usepackage{lineno}
\usepackage{threeparttable}
\usepackage{hyperref}
\usepackage{colortbl}
\begin{document}

\title{Seeing Sarcasm Through Different Eyes: Analyzing Multimodal Sarcasm Perception in Large Vision-Language Models}

\author{Junjie Chen,~\IEEEmembership{Student Member,~IEEE}, Xuyang Liu, Subin Huang, Linfeng Zhang, Hang Yu,~\IEEEmembership{Member,~IEEE}
\thanks{Junjie Chen and Subin Huang~(Corresponding Author) are with Anhui Polytechnic University~(AHPU), China~(emails:~chenjunjie@stu.ahpu.edu.cn,~subinhuang@ahpu.edu.cn).}
\thanks{Hang Yu is with Shanghai University~(SHU), China~(email: yuhang@shu.edu.cn).}
\thanks{Linfeng Zhang is with Shanghai Jiao Tong University~(SJTU), China~(email: zhanglinfeng@sjtu.edu.cn).}
\thanks{Xuyang Liu is with Sichuan University~(SCU), China~(email: liuxuyang@stu.scu.edu.cn).}
}

\markboth{Journal of \LaTeX\ Class Files,~Vol.~14, No.~8, August~2021}%
{Shell \MakeLowercase{\textit{et al.}}: A Sample Article Using IEEEtran.cls for IEEE Journals}


\maketitle

\begin{abstract}
With the advent of large vision-language models (LVLMs) demonstrating increasingly human-like abilities, a pivotal question emerges: do different LVLMs interpret multimodal sarcasm differently, and can a single model grasp sarcasm from multiple perspectives like humans?
To explore this, we introduce an analytical framework using systematically designed prompts on existing multimodal sarcasm datasets.
Evaluating 12 state-of-the-art LVLMs over 2,409 samples, we examine interpretive variations within and across models, focusing on confidence levels, alignment with dataset labels, and recognition of ambiguous ``neutral'' cases.
We further validate our findings on a diverse 100-sample mini-benchmark, incorporating multiple datasets, expanded prompt variants, and representative commercial LVLMs.
Our findings reveal notable discrepancies---across LVLMs and within the same model under varied prompts.
While classification-oriented prompts yield higher internal consistency,
models diverge markedly when tasked with interpretive reasoning.
These results challenge binary labeling paradigms by highlighting sarcasm's subjectivity.
We advocate moving beyond rigid annotation schemes toward multi-perspective,
uncertainty-aware modeling,
offering deeper insights into multimodal sarcasm comprehension.
Our code and data are available at:  
\url{https://github.com/CoderChen01/LVLMSarcasmAnalysis}.
\end{abstract}


\section{Introduction}
\IEEEPARstart{S}{arcasm}---often viewed as a nuanced form of irony---arises from a discrepancy between the literal expression and the intended implication, rendering it a persistent challenge in computational perception.
This complexity becomes even more pronounced in multimodal contexts, where textual and visual cues converge to produce rich yet ambiguous signals \cite{joshi2017automatic, cai2019multi}.
Current research on multimodal sarcasm primarily focuses on three tasks:
(1) \textit{Detection}, which categorizes text-image pairs as sarcastic or non-sarcastic~\cite{cai2019multi,qin2023mmsd2,chen2024interclipmep};
(2) \textit{Explanation}, which elucidates the rationale for labeling a text-image pair as sarcastic~\cite{desaiNicePerfumeHow2022};
and (3) \textit{Localization}, which pinpoints the specific elements in text-image pairs that trigger sarcasm~\cite{wang-etal-2022-multimodal,Singh2024WellNW}.
These lines of inquiry predominantly rely on specialized datasets such as MMSD~\cite{cai2019multi} and MMSD2.0~\cite{qin2023mmsd2},
which impose binary (sarcastic vs. non-sarcastic) or similarly rigid labels.
However, from a cognitive standpoint, human interpretations of sarcasm vary considerably across individuals~\cite{camp2012sarcasm}.
Consequently, any given text-image pair may plausibly be interpreted both as sarcastic and non-sarcastic,
calling into question the objectivity of definitive ground truth labels.
This underscores the necessity of building systems capable of reasoning from multiple interpretive angles---a capacity that demands human-like cognitive reasoning rather than mere adaptation to annotated data.

Recent advances in large vision-language models (LVLMs) such as LLaVA~\cite{liu2023visual}, Qwen-VL~\cite{bai_qwen-vl_2023}, and InternVL~\cite{chen2024far} offer a promising avenue for achieving this goal, as they exhibit increasingly human-like abilities to handle multimodal content.
\begin{figure}[t]
\centering
\includegraphics[width=\linewidth]{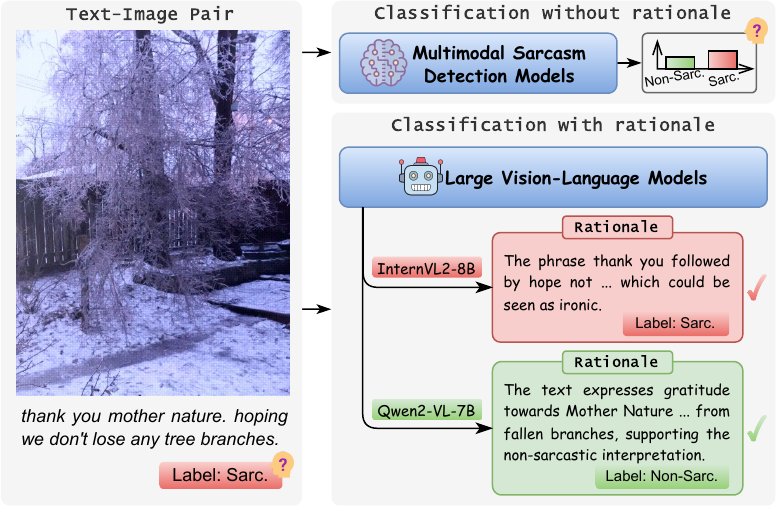}
\caption{
An example illustrating the limitations of traditional multimodal sarcasm detection models, which classify text-image pairs without rationale, compared to LVLMs, which offer explicit justifications and diverse interpretative perspectives.
}
\label{fig:motivation}
\end{figure}
As illustrated in Figure~\ref{fig:motivation}, when confronted with a text-image pair labeled \emph{sarcastic}, conventional multimodal sarcasm detection systems typically yield a binary label without any accompanying rationale.
By contrast, LVLMs are equipped not only to render an explicit classification but also to furnish multifaceted justifications, sometimes exhibiting divergent interpretations that mirror human cognitive processes.
For instance, \texttt{InternVL2-8B} confidently deems the sample sarcastic, citing overtly ironic gratitude, whereas \texttt{Qwen2-VL-7B} arrives at a non-sarcastic reading, emphasizing the literal physical dangers of icy conditions.

This example prompts a vital question: \emph{Do different LVLMs interpret sarcasm in varying ways, and can individual models demonstrate a human-like capacity to parse both sarcastic and literal meanings?}
To address this question systematically, we propose a novel four-task evaluation framework: \textbf{(1) Binary Sarcasm Classification (BSC)}, \textbf{(2) Ternary Sarcasm Classification (TSC)}, \textbf{(3) Sarcasm-Centric Scoring (SCS)}, and \textbf{(4) Literal-Centric Scoring (LCS)}.
Specifically, BSC demands that an LVLM categorize a sample as sarcastic or non-sarcastic while articulating the reasoning behind its verdict; TSC extends this to a \emph{neutral} category, reflecting the possibility that a text-image pair can be seen as either sarcastic or non-sarcastic.
Meanwhile, SCS directs a model to interpret a sample specifically through a sarcastic lens (with confidence scores), whereas LCS confines its reading to a purely literal perspective (also scored).
We implement prompt variations within each of these four tasks to probe internal consistency and compare aggregate results (derived via voting across prompts) against nominal ground truth labels.
Furthermore, qualitative assessments reveal the limitations of conventional binary labeling in capturing the inherently subjective nature of sarcasm, and an analysis of confidence distributions illuminates how models weigh neutral or ambiguous samples.

Through comprehensive experiments on 12 cutting-edge LVLMs over 2,409 multimodal samples from the MMSD2.0 dataset~\cite{qin2023mmsd2}, we observe notable variability among models as well as within individual models' responses to different prompts.
Notably, smaller models are more susceptible to prompt variations, whereas larger models display relatively stable performance yet still show biases toward sarcastic or literal interpretations.
Moreover, explicit prompting for neutral judgments exposes considerable ambiguity beyond traditional binary labels.
These findings highlight LVLMs' potential for rich, human-like multimodal sarcasm comprehension while simultaneously revealing the constraints of current approaches, which rely heavily on datasets that reduce sarcasm to a rigid binary category.
This calls for a paradigm shift toward uncertainty-aware, multi-perspective modeling methods that account for the subjective nuances of sarcasm.
To test the robustness and generalizability of our conclusions, we further construct a diverse 100-sample mini-benchmark that integrates multiple datasets, expanded prompt variants, and representative commercial LVLMs.
This supplementary analysis is complemented by human evaluation, which confirms that even state-of-the-art models produce plausible but interpretively variable outputs under different prompting conditions.

In summary, our chief contributions are:
\begin{itemize}
\item We propose a new evaluation framework comprising four carefully crafted tasks, enabling a multi-perspective assessment of LVLMs for multimodal sarcasm.
\item We demonstrate how different LVLMs diverge significantly in their sarcasm assessments, exposing the pitfalls of over-reliance on binary ground truth annotations.
\item We present empirical evidence for an uncertainty-aware, multi-perspective modeling paradigm, offering a promising direction for multimodal sarcasm research that aligns more closely with the complexities of human cognition.
\end{itemize}

\section{Related Work}
\subsection{Multimodal Sarcasm Detection}
Effective sarcasm detection plays a crucial role in tasks like sentiment analysis and opinion mining \cite{prasad2017sentiment,kannangara2018mining,SINCEREAxel2024,10521711}, making it a widely researched topic.
Early studies focused predominantly on sarcasm detection within the textual modality \cite{joshi2017automatic,SarcasmNirmala2024}.
With the rise of social multimedia platforms such as Twitter and Rbbit, multimodal sarcasm detection (MSD) based on text-image pairs has gained significant attention~\cite{ijcai2024p887}.
Cai et al.~\cite{cai2019multi} pioneered a hierarchical fusion-based model, integrating image, text, and image attribute features, and introduced a benchmark to evaluate its performance.
Subsequently, models leveraging GNNs, OCR, object detection, and transformer-based fusion techniques have emerged \cite{xu2020reasoning,pan-etal-2020-modeling,liang2021multi,Liang2022MultiModalSD,Liu2022TowardsMS,Tian2023DynamicRT,wen2023dip,wei2024g2sam,Jia_Xie_Jing_2024}.

Notably, Qin et al.~\cite{qin2023mmsd2} demonstrated that existing MSD models over-rely on spurious cues in text, resulting in an overestimation of their true multimodal sarcasm understanding capabilities.
They introduced a more robust benchmark, MMSD2.0, for evaluating these models and proposed a Multi-view CLIP model based on CLIP's vision-language representations to enhance sarcasm comprehension.
Building on this foundation, more sophisticated models have been proposed~\cite{chen2024interclipmep,Zhu2024TFCDTM,wang2024rclmufn,zhu-etal-2024-dglf,guo-etal-2025-multi}, leveraging enhanced multimodal interactions and efficient fine-tuning strategies for downstream tasks~\cite{m2istliu2025}.
With the rising prominence of large vision-language models (LVLMs) like LLaVA \cite{liu2023visual}, Tang et al.~\cite{tang-etal-2024-leveraging} developed a demonstration-based matcher and used few-shot prompting techniques to improve MSD performance.
Similarly, Wang et al.~\cite{10.1145/3690642} introduced S\textsuperscript{3}Agent, a multi-view agent framework aimed at enhancing the zero-shot MSD capabilities of LVLMs.
In addition to research on text-image pairs, efforts to design models capable of handling more complex modalities (e.g., video and speech) have also emerged \cite{castro-etal-2019-towards,ray-etal-2022-multimodal,bhosale2023sarcasm}.

Despite the significant advancements in MSD, existing models fail to provide rationales for their predicted labels.
They overlook the inherent subjectivity of sarcasm, where the same sample might reasonably be interpreted as both sarcastic and non-sarcastic.
As a result, these models fail to identify neutral samples, which can be understood from both perspectives, challenging the validity of binary classifications.
Enhancing multimodal sarcasm classification with rationale generation and identifying neutral samples is crucial for developing reliable sarcasm understanding systems that better capture the complexities of sarcastic expressions.

\subsection{Beyond Multimodal Sarcasm Detection}
Beyond MSD, another line of research focuses on more advanced sarcasm understanding tasks, such as multimodal sarcasm explanation, target identification, and reasoning \cite{desaiNicePerfumeHow2022,kumar-etal-2022-become,wang-etal-2022-multimodal,jing-etal-2023-multi,Singh2024WellNW,Du2023DocMSUAC,chen2024cofipara}.
Desai et al.~\cite{desaiNicePerfumeHow2022} were the first to propose the task of multimodal sarcasm explanation, aimed at generating natural language descriptions to explain the sarcastic semantics of given text-image pairs.
Similarly, Wang et al.~\cite{wang-etal-2022-multimodal} introduced the task of multimodal sarcasm target identification, which seeks to extract the specific words and objects within text-image pairs that trigger sarcasm.
Building further, Singh et al.~\cite{Singh2024WellNW} investigated sarcasm detection within multimodal conversations, identifying its origins and providing rationales for the detected sarcastic semantics. They coined this task ``Sarcasm Initiation and Reasoning in Conversations (SIRC).''

Although the above studies transcend binary labels and delve into multidimensional sarcasm understanding, they still overlook the subjectivity inherent in sarcasm and fail to address the challenges posed by neutral samples.
Moreover, these studies are limited by their reliance on small datasets for training and evaluation, raising concerns about model generalizability.

\subsection{Prompting Engineering for Sarcasm Understanding}
Text-based sarcasm detection has been extensively studied, with recent research leveraging prompt engineering to assess large language models' capacity for understanding sarcasm~\cite{yao2024sarcasmdetectionstepbystepreasoning,zhang2024sarcasmbench,lee2024pragmaticmetacognitivepromptingimproves}.
However, the emergence of multimodal models—such as closed-source LVLMs like GPT-4o~\cite{openai2024gpt4o} and open-source models like LLaVA~\cite{liu2023visual}—which demonstrate human-like reasoning across both textual and visual modalities, raises the question of whether these models can effectively interpret sarcasm in a multimodal setting.
Moreover, existing text-based studies primarily emphasize binary sarcasm classification, often neglecting the need for rationales and overlooking the inherent subjectivity of sarcasm.
As a result, they offer a constrained evaluation of a model's ability to truly comprehend and interpret sarcastic language.

To bridge these gaps, we propose a novel evaluation framework that prompts LVLMs to identify sarcasm, explicitly recognize neutral samples, and assess examples from both sarcastic and non-sarcastic perspectives—while requiring rationales for all judgments.
Through both qualitative and quantitative analyses of model outputs, we gain insights into how different LVLMs interpret sarcasm, as well as the subjectivity embedded in sarcasm perception across and within models.

\section{Methodology}
\begin{figure*}[t]
\centering
\includegraphics[width=0.97\linewidth]{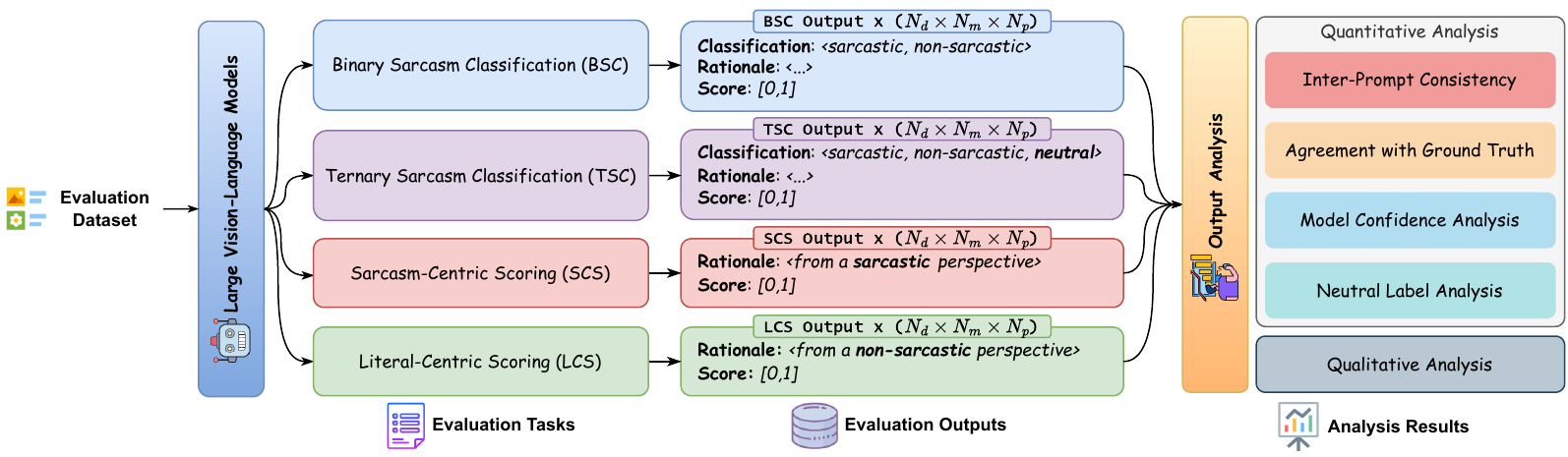}
\caption{
Overview of the evaluation framework.
Our framework takes as input an evaluation dataset consisting of \( N_d \) text-image pairs with predefined labels and prompts \( N_m \) LVLMs to perform four distinct tasks, each with \( N_p \) prompt variants.
This results in \( 4 \times N_d \times N_m \times N_p \) evaluation outputs, which are systematically analyzed through quantitative and qualitative assessments.
}
\label{fig:framework-overview}
\end{figure*}
In this section, we introduce our evaluation framework for multimodal sarcasm understanding. As shown in Figure~\ref{fig:framework-overview}, our approach consists of four complementary tasks---Binary Sarcasm Classification (BSC), Ternary Sarcasm Classification (TSC), Sarcasm-Centric Scoring (SCS), and Literal-Centric Scoring (LCS)---each designed to assess LVLMs from different perspectives.
These tasks generate classification decisions, rationales, and confidence scores, which are further analyzed through both quantitative and qualitative evaluations to examine inter-prompt consistency, agreement with ground truth, model confidence, and neutral sample recognition.
The following subsections define each task and describe the prompt designs.

\subsection{Task Definition}\label{sec:task-definition}
We introduce four distinct tasks designed to assess LVLMs' capability for multimodal sarcasm understanding from multiple perspectives, each described in detail below.

\textbf{Binary Sarcasm Classification (BSC): }
The BSC task requires the model to classify a given text-image pair \((T_i, V_i)\) into \textit{sarcastic} or \textit{non-sarcastic} categories. It is defined as:
\begin{equation}
\mathbf{\pi}_{\textrm{LVLM}} \left( \mathcal{P}^{\textrm{BSC}}\left( T_i, V_i \right) \right) = \left\{ \hat{Y}^{\textrm{BSC}}_i, R^{\textrm{BSC}}_i, S^{\textrm{BSC}}_i \right\},
\end{equation}
where \(\mathcal{P}^{\textrm{BSC}}(\cdot)\) denotes the prompt template specifying the task, and \(\hat{Y}^{\textrm{BSC}}_i \in \{\textrm{sarcastic}, \textrm{non-sarcastic}\}\) represents the predicted label.
Additionally, \(R^{\textrm{BSC}}_i\) is the rationale for the prediction, and \(S^{\textrm{BSC}}_i \in [0, 1]\) is the confidence score.
The inclusion of \( R^{\textrm{BSC}}_i \) provides interpretability, allowing for a deeper analysis of the model's reasoning process and potential biases in sarcasm detection.

\textbf{Ternary Sarcasm Classification (TSC): }
To test whether LVLMs can directly identify \textit{neutral} samples, we introduce the TSC task, defined as:
\begin{equation}
\mathbf{\pi}_{\textrm{LVLM}} \left( \mathcal{P}^{\textrm{TSC}}\left( T_i, V_i \right) \right) = \left\{ \hat{Y}^{\textrm{TSC}}_i, R^{\textrm{TSC}}_i, S^{\textrm{TSC}}_i \right\},
\end{equation}
where \(\mathcal{P}^{\textrm{TSC}}(\cdot)\) is similar to \(\mathcal{P}^{\textrm{BSC}}(\cdot)\), except that it enables the model to output ternary labels:
\(\hat{Y}^{\textrm{TSC}}_i \in \{\textrm{sarcastic}, \textrm{non-sarcastic}, \textrm{neutral}\}\).
The inclusion of the \textit{neutral} category enables a more nuanced assessment of sarcasm perception and ambiguity recognition in LVLMs.

\textbf{Sarcasm-Centric Scoring (SCS): }
This task requires the model to score the degree to which a given text-image pair can be interpreted as \textit{sarcastic}:
\begin{equation}
\mathbf{\pi}_{\textrm{LVLM}} \left( \mathcal{P}^{\textrm{SCS}}\left( T_i, V_i \right) \right) = \left\{ R^{\textrm{SCS}}_i, S^{\textrm{SCS}}_i \right\},
\end{equation}
where \(\mathcal{P}^{\textrm{SCS}}(\cdot)\) is the prompt template, and \(R^{\textrm{SCS}}_i\) and \(S^{\textrm{SCS}}_i\) denote the rationale and sarcasm-centric score, respectively. A score closer to 0 indicates a weak sarcastic interpretation, while a score closer to 1 indicates strong agreement.
For the purpose of classification-based analysis, we derive sarcasm labels from the SCS task as follows:
\begin{equation}
\hat{Y}^{\textrm{SCS}}_i =
\begin{cases}
\textrm{sarcastic}, & \text{if } S^{\textrm{SCS}}_i > 0.5 \\
\textrm{non-sarcastic}, & \text{otherwise}.
\end{cases}
\end{equation}

\textbf{Literal-Centric Scoring (LCS): }
Similar to SCS, this task requires the model to score how well a given text-image pair can be interpreted from a literal, \textit{non-sarcastic} perspective:
\begin{equation}
\mathbf{\pi}_{\textrm{LVLM}} \left( \mathcal{P}^{\textrm{LCS}}\left( T_i, V_i \right) \right) = \left\{ R^{\textrm{LCS}}_i, S^{\textrm{LCS}}_i \right\}.
\end{equation}
Here, \(\mathcal{P}^{\textrm{LCS}}(\cdot)\) is the prompt template for literal-centric reasoning, and \(R^{\textrm{LCS}}_i\) and \(S^{\textrm{LCS}}_i\) denote the rationale and literal-centric score, respectively. A score closer to 0 indicates weak literal interpretation, while a score closer to 1 indicates strong agreement.
For classification-based analysis, we derive sarcasm labels from the LCS task as follows:
\begin{equation}
\hat{Y}^{\textrm{LCS}}_i =
\begin{cases}
\textrm{sarcastic}, & \text{if } S^{\textrm{LCS}}_i < 0.5 \\
\textrm{non-sarcastic}, & \text{otherwise}.
\end{cases}
\end{equation}

\subsection{Prompt Design}
\begin{figure}
\centering
\includegraphics[width=\linewidth]{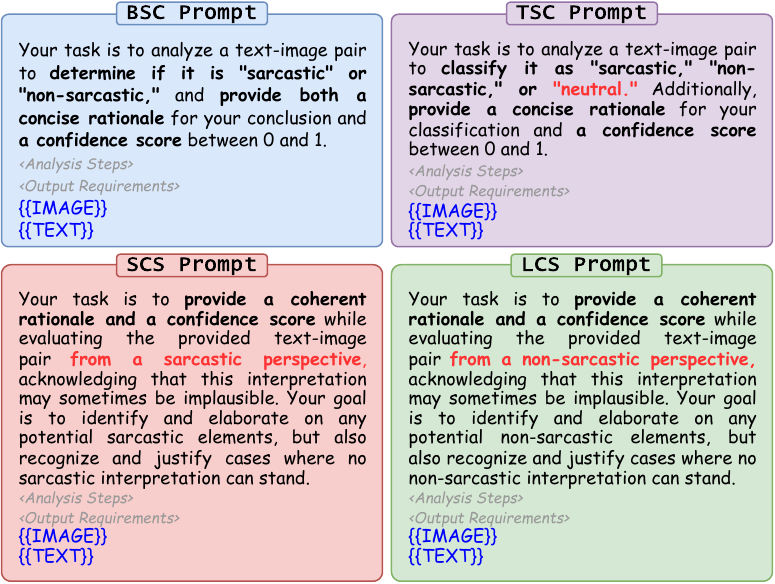}
\caption{
Example prompts for each task.
For each task prompt, we first provide a \textit{task description}, followed by explicit instructions for the \textit{analysis steps}, and conclude by specifying \textit{the required output format}.
In the figure, the descriptions of analysis steps and output requirements are omitted for brevity.
Detailed prompts can be found in Appendix~\ref{app:prompt-details}.
\textcolor{blue}{\{\{IMAGE\}\}} and \textcolor{blue}{\{\{TEXT\}\}} represent the input image and text, respectively.
}
\label{fig:prompt-demo}
\end{figure}
For \( \mathcal{P}^{\textrm{BSC}}, \mathcal{P}^{\textrm{TSC}}, \mathcal{P}^{\textrm{SCS}}, \) and \( \mathcal{P}^{\textrm{LCS}} \), we adopt a zero-shot prompting approach.
Inspired by the Plan-and-Solve prompting method \cite{wang-etal-2023-plan} and the unique characteristics of multimodal sarcasm, each task prompt is structured as follows: we first provide a clear \textit{description of the task}, followed by explicitly specifying the \textit{analytical steps} to guide the model toward stronger reasoning.
Finally, the model is prompted to output in a \textit{predefined format}.
As illustrated in Figure~\ref{fig:prompt-demo}, we provide examples of prompts for the four tasks. Detailed prompts for each task are available in Appendix~\ref{app:prompt-details}.

Specifically, since the outputs of LVLMs have been shown to be highly sensitive to variations in task prompts~\cite{dumpala2024sensitivity,mizrahi2024state}, we design \( N_p \) prompt variants for each task.
These variants differ only at the surface level (e.g., wording or phrasing) while maintaining the same task description, analytical steps, and output requirements.
This strategy ensures greater robustness and reliability of the conclusions drawn from the analysis.
Therefore, given a dataset \( \mathcal{D}=\left\{(T_i,V_i)\right\}_{i=1}^{N_d} \) and \( N_m \) selected LVLMs, the execution of each task yields \( N_d \times N_p \times N_m \) results.

\section{Experimental Setup}
In this section, we provide a detailed overview of the evaluation dataset used in our experiments, introduce the selected LVLMs, and describe the implementation details of the evaluation framework, including key methodologies and procedures.

\paragraph{Evaluation Dataset}
The MMSD2.0~\cite{qin2023mmsd2} test set, consisting of 2,409 text-image pairs, i.e., \( N_d = 2409 \), is used as the evaluation dataset.
This dataset includes 1,037 sarcastic samples and 1,372 non-sarcastic samples, providing a balanced and comprehensive benchmark for sarcasm detection. 
Concretely, MMSD2.0 is an improved version of the original dataset proposed by Cai et al.~\cite{cai2019multi}, with significant enhancements that include the removal of spurious cues in the textual data and the correction of mislabeled samples.
This dataset has since been widely adopted in recent research on multimodal sarcasm detection~\cite{chen2024interclipmep,Zhu2024TFCDTM,wang2024rclmufn,zhu-etal-2024-dglf,guo-etal-2025-multi}.
Thus, its use as the test dataset ensures the evaluation is performed on a high-quality, reliable benchmark with minimized noise.
This enables a robust assessment of LVLMs in understanding multimodal sarcasm.
The cleaned and corrected annotations provide a strong foundation for fair comparisons and meaningful analysis.

\begin{table}[t]
\centering
\caption{
Details of selected LVLMs.
}
\label{tab:LVLM-details}
\begin{threeparttable}
\begin{tabularx}{\linewidth}{Xc}
\toprule
\textbf{Model Name (Short Name)}                      & \textbf{Architecture}                                       \\
\midrule
HuggingFaceTB/SmolLVLM-Instruct~(\textbf{smo-2B})               & Idefics3~\cite{laurençon2024building}                      \\
OpenGVLab/InternVL2-2B~(\textbf{int2-2B})                      & InternVL2~\cite{chen2024far}                                \\
Qwen/Qwen2-VL-2B-Instruct~(\textbf{qw2-2B})                    & Qwen2VL~\cite{Qwen2VL}                                     \\
microsoft/Phi-3.5-vision-instruct~(\textbf{phi3.5-4B})         & Phi3~\cite{abdin2024phi3}                                  \\
llava-hf/llava-1.5-7b-hf~(\textbf{lav1.5-7B})                  & LLaVA-1.5~\cite{liu2024improvedbaselinesvisualinstruction} \\
llava-hf/llava-v1.6-mistral-7b-hf~(\textbf{lav1.6-7B})         & LLaVA-Next~\cite{liu2024llavanext}                         \\
Qwen/Qwen2-VL-7B-Instruct~(\textbf{qw2-7B})                    & Qwen2VL~\cite{Qwen2VL}                                     \\
OpenGVLab/InternVL2-8B~(\textbf{int2-8B})                      & InternVL2~\cite{chen2024far}                                \\
meta-llama/Llama-3.2-11B-Vision~(\textbf{lam3.2-11B})          & Llama-3.2~\cite{llama32}                                    \\
llava-hf/llava-1.5-13b-hf~(\textbf{lav1.5-13B})                & LLaVA-1.5~\cite{liu2024improvedbaselinesvisualinstruction} \\
Qwen/Qwen2-VL-72B-Instruct~(\textbf{qw2-72B})                  & Qwen2VL~\cite{Qwen2VL}                                     \\
OpenGVLab/InternVL2-Llama3-76B~(\textbf{int2-76B})             & InternVL2~\cite{chen2024far}                                \\
\bottomrule
\end{tabularx}
\begin{tablenotes}
\footnotesize
\item[*] In this table, we list the full names of the selected LVLMs from the HuggingFace model hub~\cite{wolf2019huggingfaces}, along with their respective architectures.
To facilitate the presentation of subsequent experimental results, we have assigned each model a short name, following the convention: \{abbreviation\}-\{model size\}.
\end{tablenotes}
\end{threeparttable}
\end{table}

\paragraph{Selected LVLMs}
To comprehensively evaluate the performance of existing LVLMs on our proposed task, we select a diverse set of open-source models with varying architectures and scales, totaling 12 models, i.e., \( N_m = 12 \).  
Our selection includes Idefics3~\cite{laurençon2024building}, InternVL 2.0~\cite{chen2024far}, Qwen2-VL~\cite{Qwen2VL}, Phi3~\cite{abdin2024phi3}, LLaVA-1.5~\cite{liu2024improvedbaselinesvisualinstruction}, LLaVA-NeXT~\cite{liu2024llavanext}, and Llama-3.2~\cite{llama32}.  
For detailed specifications of the selected LVLMs, please refer to Table~ \ref{tab:LVLM-details}.

\paragraph{Implementation Details}
We run all models using the OpenAI Compatible Server feature provided by the vLLM framework~\cite{kwon2023efficient}, and all requests are sent via the OpenAI client library~\footnote{\url{https://github.com/openai/openai-python}}.
For each task-specific prompt, we design three variants, i.e., \( N_p = 3 \).
Details can be found in Appendix~\ref{app:prompt-details}.
To maintain determinism and evaluate the intrinsic capabilities of each model, we use greedy decoding as the default generation strategy.

However, we observe that certain models generate rationales exceeding the context length limit.
To mitigate this, we introduce a soft constraint in our prompts, specifying \textit{``limit to \(\{x\}\) words,''} with \( x \) set to 150 by default in our experiments. Despite this constraint, some models still fail to comply.
In such cases, we adopt an adaptive adjustment strategy. If a model exceeds the length limit, we decrement \( x \) by 10 iteratively until a valid response is obtained.
If \( x \) reaches 0, we abandon greedy decoding and instead set a fixed random seed of 42 while setting the temperature to 0.1.
If the issue persists, we gradually increase the temperature by 0.1 until a valid output is generated.
If the temperature reaches 1.0 and failures persist, we keep the temperature fixed and increase the random seed by increments of 10 until a valid output is obtained.
By employing this strategy, we maximize reproducibility while ensuring the successful generation of rationales within the expected constraints. 

\section{Inter-Prompt Consistency}
In this section, we analyze the consistency of model responses across semantically equivalent prompts for each task, examining whether variability in predictions stems from sensitivity to prompt phrasing rather than genuine multimodal understanding.
To assess this, the following subsections introduce the metrics and provide a detailed analysis of inter-prompt consistency across the four tasks.

\subsection{Classification Consistency}
\subsubsection{Metric}
As described in Section~\ref{sec:task-definition}, each task yields predicted classification labels, enabling a classification-based qualitative analysis.
To quantify consistency, we use Krippendorff's $\alpha$~\cite{krippendorff2004reliability} as the Classification Consistency Score, as it effectively handles partially missing or unavailable data—particularly relevant in cases where models fail to follow instructions, resulting in missing predictions for certain samples. However, a limitation of this metric is its potential instability in scenarios with extreme class distributions or small datasets.
Krippendorff's $\alpha$ has a range of $[-1,1]$, where a value closer to 1 indicates higher consistency, and a value closer to 0 indicates poorer consistency.
Negative values generally indicate that the variance in the ratings between evaluators is greater than that expected by random consistency.

\subsubsection{Result}
\begin{figure}
\centering
\includegraphics[width=\linewidth]{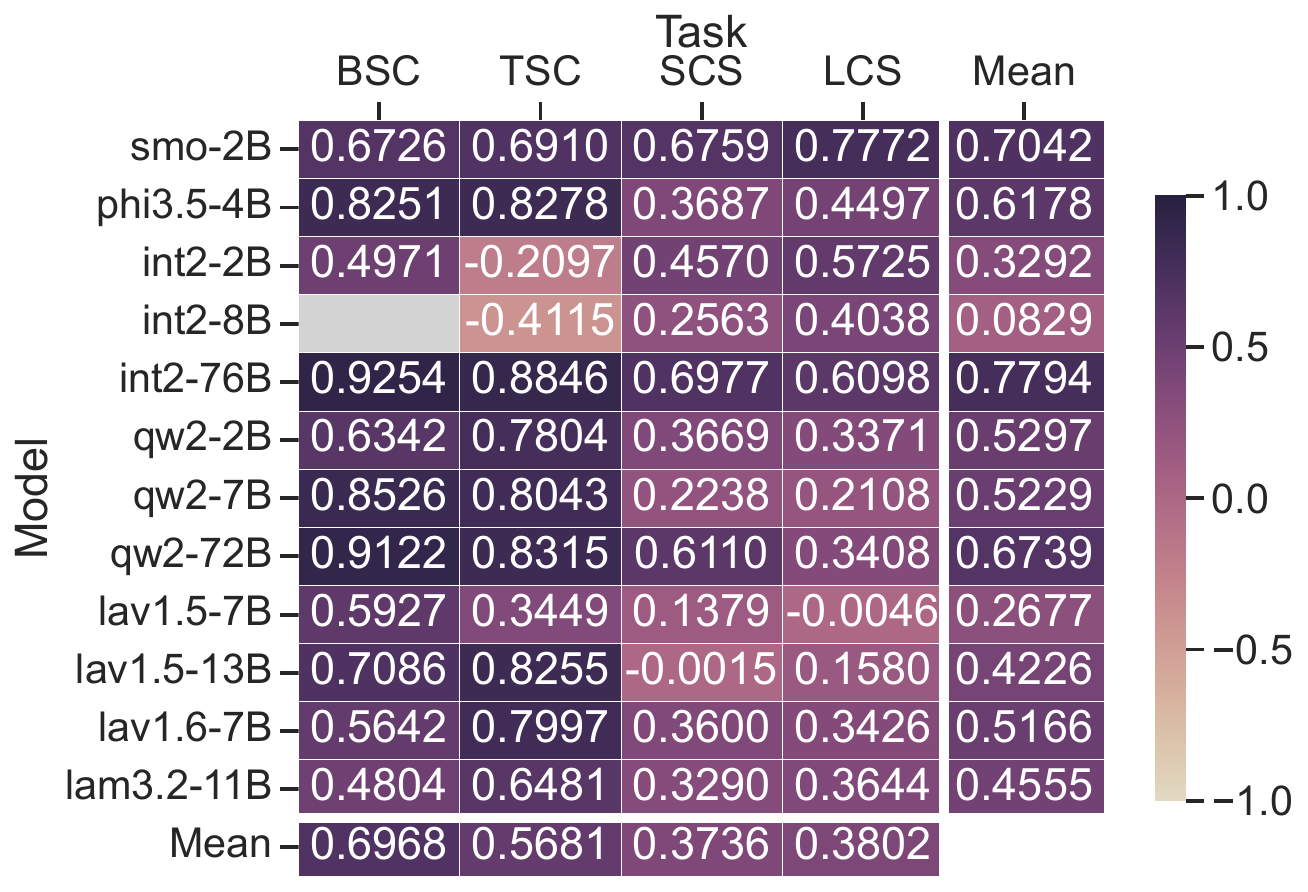}
\caption{
Classification consistency heatmap.
This figure illustrates the consistency scores of classification results from different prompt variants across four tasks for each model.
Specifically, the gray blocks indicate that the Krippendorff’s $\alpha$ calculation for \texttt{int2-8B} in the BSC task is invalid.
Further inspection reveals that \texttt{int2-8B} classifies all samples as \textit{sarcastic} across prompt variants, preventing the computation of Krippendorff’s $\alpha$.
}
\label{fig:cls-consistency}
\end{figure}

As shown in Figure~\ref{fig:cls-consistency}, models show higher consistency scores on the BSC and TSC tasks compared to the SCS and LCS tasks.
This suggests that models perform better when directly classifying sarcasm (BSC) or distinguishing between sarcastic and non-sarcastic samples with a neutral option (TSC).
These tasks are more straightforward, involving clear-cut classification that allows models to focus on simpler patterns.
In contrast, the SCS and LCS tasks require more nuanced reasoning, as models need to interpret samples from sarcasm-centric (SCS) or literal-centric (LCS) perspectives.
This additional cognitive complexity likely results in lower consistency scores, as models face greater challenges in handling the subjective and multifaceted nature of sarcasm.
From the model perspective, we observe that models with fewer than 13B parameters are more sensitive to prompt variations than those with over 70B parameters.
This suggests that larger models, with greater capacity, may be better equipped to generalize and are less reliant on prompt phrasing, allowing them to navigate the inherent subjectivity in sarcasm detection more effectively.
This results in more stable performance across different prompt variants.
Nevertheless, there are exceptions.
Models like \texttt{smo-2B} and \texttt{phi3.5-4B}, which are smaller in scale, still show relatively high consistency.
This indicates that while model size does influence robustness, smaller models can still achieve competitive performance, possibly due to their specific architectural optimizations or the way they handle task-specific patterns.

Furthermore, models based on the QwenVL architecture consistently outperform others, regardless of model size.
This suggests that the QwenVL architecture may have inherent advantages in handling sarcasm detection tasks, potentially due to its design choices, such as better integration of multimodal information or more robust processing of subtle linguistic cues.
These findings indicate that architecture may play a critical role in performance consistency, beyond just model size.

\subsection{Rationale Consistency}
\subsubsection{Metric}
For all tasks, we require the model to provide a rationale corresponding to its classification or score.
Thus, we also explore the consistency of rationale outputs under different prompt variants.
For each sample across different prompt variants, we calculate the BERTScore~\cite{Zhang2020BERTScore} F1 measure for the pairwise rationale texts.
Specifically, each pair must have the same classification label; otherwise, the pair is discarded.
Finally, we compute the average of all the scores as the Rationale Consistency Score. Notably, this score ranges from 0 to 1, with a score closer to 1 indicating higher consistency.

\subsubsection{Result}
\begin{figure}
\centering
\includegraphics[width=0.72\linewidth]{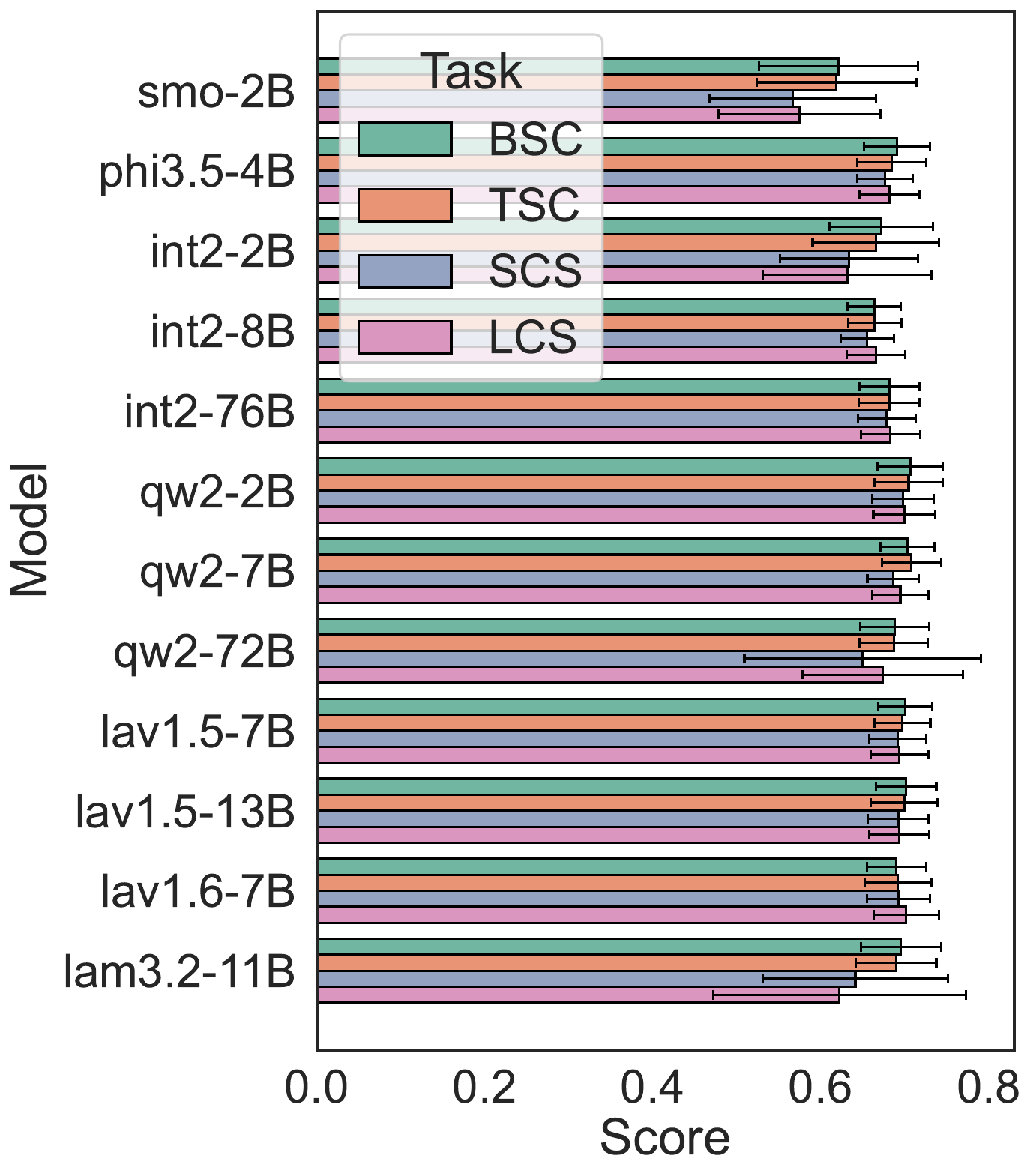}
\caption{
Rationale consistency score.
This figure displays the average similarity of rationales for the same classification across all samples, tasks, and models, where different prompt variants were used. 
}
\label{fig:rationale-consistency}
\end{figure}

As shown in Figure~\ref{fig:rationale-consistency}, similar to the classification consistency results, most models exhibit lower rationale consistency scores for the SCS and LCS tasks compared to the BSC and TSC tasks, with higher standard deviations.
This further reinforces the notion that the SCS and LCS tasks demand more nuanced reasoning and subjective interpretation.
The greater variability in rationale consistency across these tasks suggests that models are grappling with the complexities of reasoning from a sarcasm-centric or literal-centric perspective, which leads to more diverse and less stable rationale outputs.
In contrast, tasks like BSC and TSC are more straightforward, involving clear classification and therefore more consistent rationales across prompt variants.
This indicates that when the task aligns more closely with conventional classification paradigms---where models can rely on binary or structured decision-making rather than engaging in open-ended reasoning---they tend to produce more stable and aligned rationales.

Furthermore, we observe that all models tend to fluctuate within a rationale consistency range of approximately 0.6 to 0.8.
This suggests that while models are able to generate rationales that are somewhat consistent across different prompts, there is still a considerable degree of variation.
This range indicates a moderate level of stability, suggesting that while models can maintain a certain level of coherence in their reasoning, factors such as prompt phrasing, the model's interpretation of the task, and the subjective nature of sarcasm lead to some divergence in the generated rationales.

\section{Agreement with Ground Truth}\label{sec:agreement-with-ground-truth}
In this section, we first introduce our approach to aggregating the results obtained from multiple prompt variants for each task.
Subsequently, we design evaluation metrics and analyze the agreement between the aggregated model predictions and the ground truth labels of the test dataset.

\subsection{Preliminary}
As described in the previous section, for each model and each task, different prompt variants may yield different sarcasm classification labels.  
To obtain a stable classification decision, we aggregate predictions from different prompt variants using a majority voting mechanism.  
For a given model \(\mathbf{\pi}_{\textrm{LVLM}}\), task \(\mathcal{T} \in \{\textrm{BSC},\textrm{TSC},\textrm{SCS},\textrm{LCS}\}\), and sample \( (T_i, V_i) \), the final classification label \(\hat{Y}_i^{\mathcal{T}}\) is determined as:
\begin{equation}
\hat{Y}_i^{\mathcal{T}} = \arg\max_{y} \sum_{p=1}^{N_p} \mathbb{I} \left[ \hat{Y}_{i,p}^{\mathcal{T}} = y \right],
\end{equation}
where \(\hat{Y}_{i,p}^{\mathcal{T}}\) denotes the classification output for sample \( (T_i, V_i) \) using the \(p\)-th prompt, and \(\mathbb{I}[\cdot]\) is the indicator function returning 1 if the condition is met and 0 otherwise.

Beyond aggregating predictions within a single task, for the SCS and LCS tasks, classification labels can also be derived by comparing the relative magnitudes of the scores across both tasks.
Given sarcasm-centric and literal-centric scores \( S^{\textrm{SCS}}_i \) and \( S^{\textrm{LCS}}_i \), we define the classification rule as:
\begin{equation}
\hat{Y}^{\textrm{COMP}}_i =
\begin{cases}
\textrm{sarcastic}, & \text{if } S^{\textrm{SCS}}_i > S^{\textrm{LCS}}_i, \\
\textrm{non-sarcastic}, & \text{if } S^{\textrm{SCS}}_i < S^{\textrm{LCS}}_i, \\
\textrm{undefined}, & \text{otherwise}.
\end{cases}
\end{equation}

Since both SCS and LCS have \(N_p\) prompt variants, each SCS result is compared against all LCS results, leading to \(N_p \times N_p\) pairwise comparisons.
To determine the final classification label, we also apply a majority voting mechanism across all comparisons.
When recording votes, \textit{undefined} cases are ignored, meaning only comparisons where a clear sarcasm or non-sarcasm label is assigned contribute to the final decision.
\begin{figure*}
\centering
\includegraphics[width=1\linewidth]{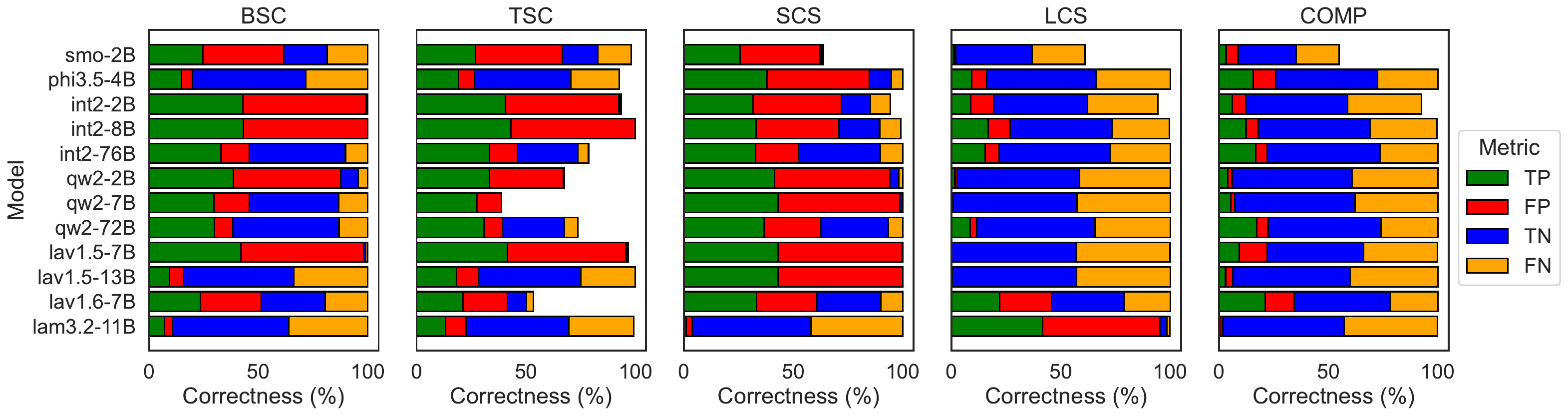}
\caption{
Classification statistics of predictions compared to ground truth.  
This figure illustrates the proportion of correctly classified samples for each method along the x-axis.  
For all samples that received a predicted classification label, we compute the distribution of true positives (TP), false positives (FP), true negatives (TN), and false negatives (FN).  
These four categories are represented in different colors, showing their respective proportions within each classification task.
}
\label{fig:stat-scores}
\end{figure*}

\subsection{Metric}  
As previously discussed, existing benchmarks provide ground truth labels only for \textit{sarcastic} and \textit{non-sarcastic} categories.
To ensure fair comparison, we restrict evaluation to samples that can be definitively classified as either sarcastic or non-sarcastic.
Concretely, the following ambiguous cases are excluded from statistical analysis:
\begin{itemize}
\item \textbf{Neutral category in the TSC task:} Given the inherently binary nature of ground truth labels, samples classified as neutral in the TSC task are excluded from statistical evaluation but are systematically analyzed separately in the neutral label analysis and qualitative study.
\item \textbf{Undefined cases in majority voting:} If there is an equal number of votes across multiple labels in the final classification, no definitive assignment can be made.
\end{itemize}

To quantify model performance, we introduce the \textit{correctness} metric, which represents the proportion of correctly classified samples among those with definitive classifications.  
For these samples, predicted labels are compared against ground truth labels, and classification statistics---including \textit{true positives (TP)}, \textit{false positives (FP)}, \textit{true negatives (TN)}, and \textit{false negatives (FN)}---are computed to assess the model's alignment with benchmark annotations.
Furthermore, to examine the overall consensus among different models, we aggregate the predictions from all models and determine the final classification label using a majority voting mechanism.  
The aggregated predictions are then evaluated against the ground truth labels to analyze the collective agreement of the models with the benchmark data.

\subsection{Result}
\begin{figure}
\centering
\includegraphics[width=0.8\linewidth]{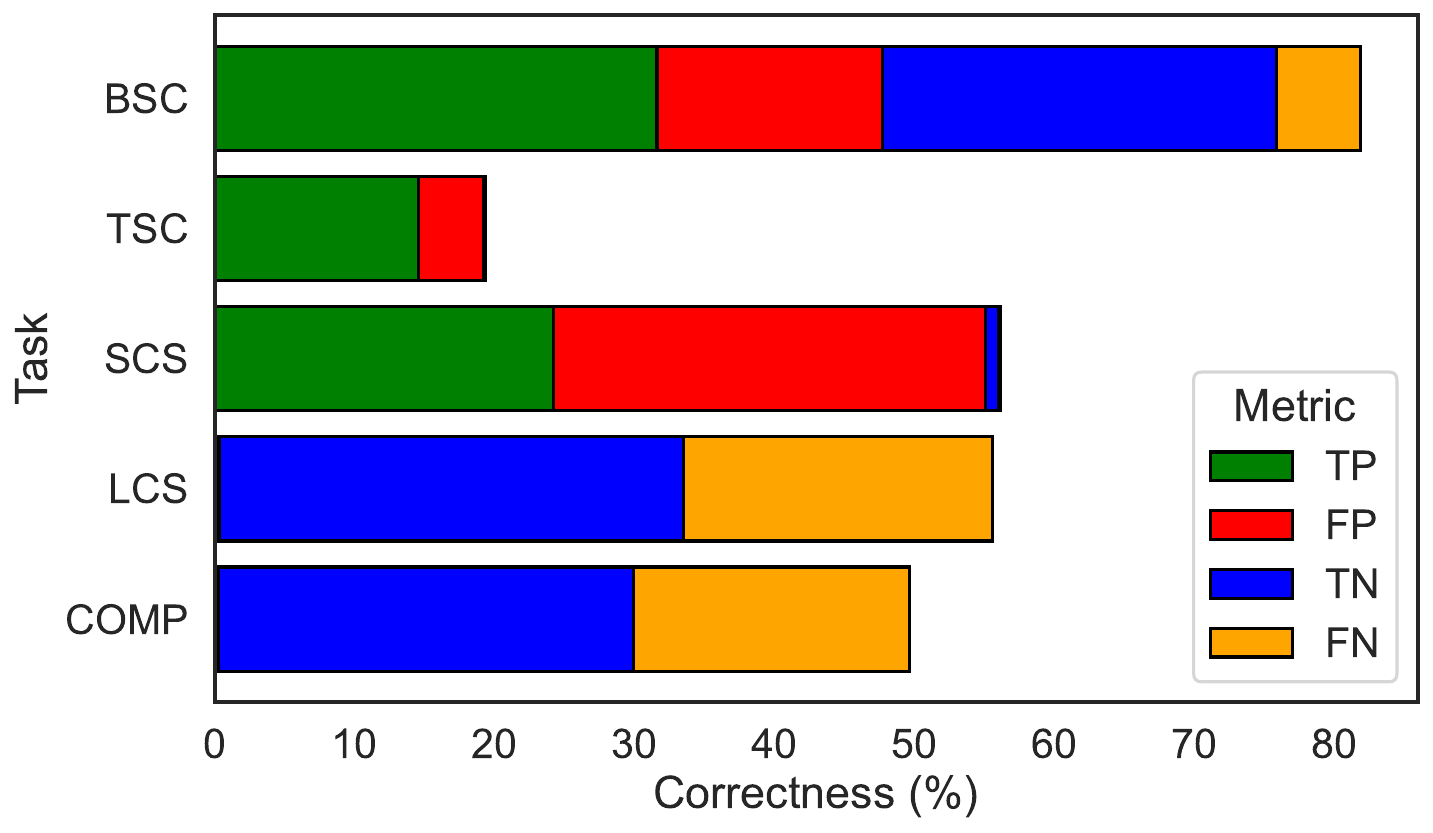}
\caption{
Classification statistics of aggregated predictions compared to ground truth. 
}
\label{fig:agg-stat-score}
\end{figure}
As shown in Figure~\ref{fig:stat-scores}, for the BSC task, most models align well with non-sarcastic ground truth, indicating an easier grasp of literal meanings.
However, models like \texttt{int2-2B} and \texttt{int2-8B} classify all samples as sarcastic, revealing a bias towards sarcasm.
Larger models, such as \texttt{int2-76B} and \texttt{qw2-72B}, maintain a balanced alignment, while smaller models tend to overfit to a single class, showing weaker sarcasm comprehension.
In TSC, adding a neutral category leads to lower correctness for some models, implying an attempt to recognize neutrality, though validity remains uncertain.
QwenVL-based models frequently classify non-sarcastic samples as neutral, demonstrating an openness to ambiguity, whereas others remain unaffected, indicating insensitivity to neutrality.
For SCS and LCS, models strongly favor their assigned perspective, showing prompt sensitivity.
Interestingly, when comparing scores in COMP, all models lean towards non-sarcastic interpretations, suggesting that literal understanding is their default.

As shown in Figure~\ref{fig:agg-stat-score}, 
aggregated results show BSC aligns best with ground truth, while TSC shifts non-sarcastic samples toward neutrality.
SCS and LCS results confirm that models are more confident in literal interpretations.

Overall, model agreement with ground truth varies.
Sarcasm detection remains challenging, with a bias toward literal meaning. This highlights the limitations of binary-labeled datasets and the need for more nuanced evaluation frameworks.
Detailed results can be found in Appendix~\ref{app:agreement-with-ground-truth-results}.

\section{Model Confidence Analysis}
\begin{figure*}[t]
\centering
\includegraphics[width=1\linewidth]{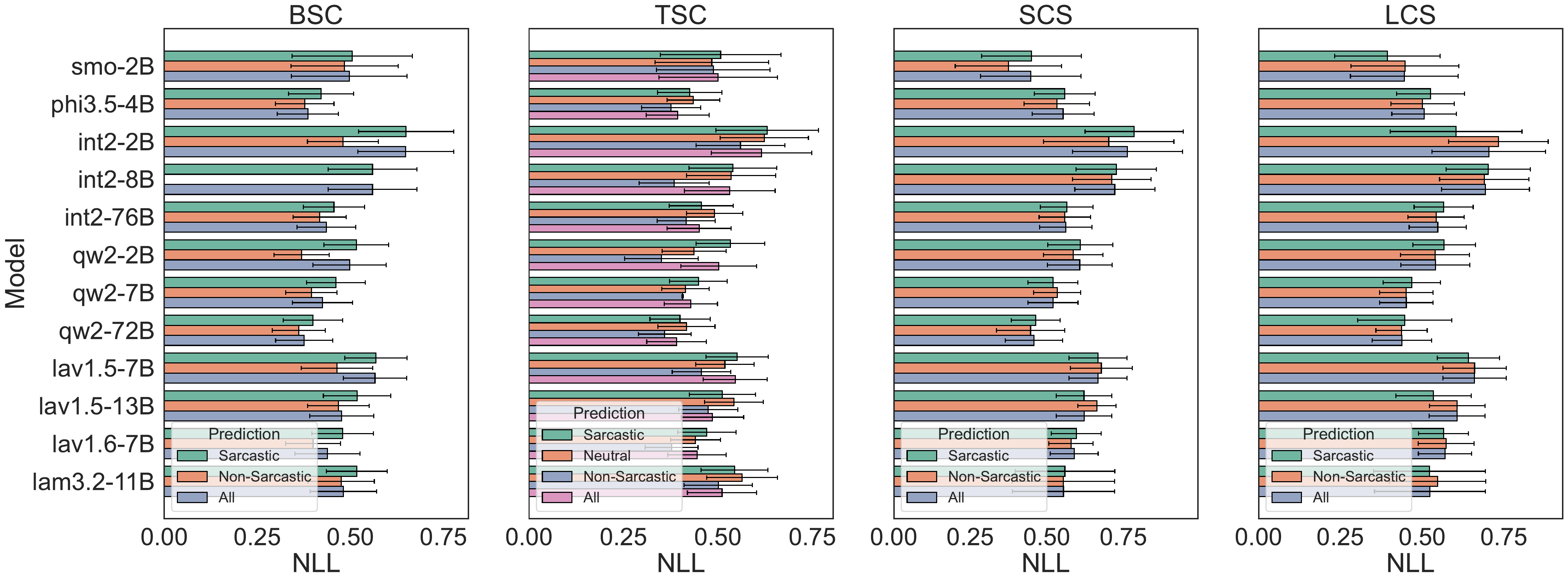}
\caption{
NLL analysis of model predictions across different tasks.
A lower NLL value indicates greater confidence.
}
\label{fig:model-nll}
\end{figure*}
In this section, we analyze how confidently models make predictions across the tasks, highlighting their varying levels of certainty in multimodal sarcasm understanding.

\subsection{Metric}
We assess model confidence using Negative Log-Likelihood (NLL), computed as the aggregated token-level log-probability of the model’s own output conditioned on the input (i.e., the image-text pair and prompt).
Lower NLL values indicate higher model certainty, while higher values suggest greater uncertainty.
NLL is a standard and theoretically grounded metric for autoregressive models, enabling consistent confidence estimation across both classification-oriented (BSC, TSC) and open-ended reasoning tasks (SCS, LCS).
In the context of subjective phenomena like sarcasm, NLL serves as a proxy for the model’s interpretive commitment, indicating how decisively it favors one perspective under a given prompt.
We emphasize that NLL is used as a within-model measure to analyze task- or prompt-level variations, rather than for direct comparison across architectures, which may differ in tokenization and decoding strategies.
Overall, NLL offers an interpretable and model-native signal for assessing confidence in multimodal sarcasm interpretation.

\subsection{Result}
As shown in Figure~\ref{fig:model-nll}, in the BSC task, most models exhibit lower confidence when predicting sarcasm compared to non-sarcasm. This suggests that sarcasm detection is inherently more challenging, possibly due to the subtle and context-dependent nature of sarcastic expressions. 
In the TSC task, the introduction of a neutral category further increases uncertainty, particularly for models that previously showed a strong bias towards either sarcastic or non-sarcastic classifications. This indicates that some models struggle to recognize the middle ground between sarcasm and literal meaning.
For SCS and LCS tasks, models display clearer trends: predictions aligned with the given interpretative prompt (sarcasm-focused for SCS, literal-focused for LCS) tend to be more confident, while opposite interpretations induce higher NLL. This reinforces the observation that prompting significantly influences model confidence, with literal understanding generally being the more stable default.
Overall, these results highlight a fundamental challenge in sarcasm detection: models exhibit greater uncertainty when identifying sarcasm than when recognizing non-sarcastic statements. This aligns with human intuition—sarcasm often relies on implicit cues, making it harder to discern with high confidence.

\section{Neutral Label Analysis}
In this section, we explore the overlap between neutral samples identified by the TSC method and those identified by comparing SCS and LCS scores.
We also analyze the proportion of sarcastic and non-sarcastic ground truth labels within neutral samples.
\begin{figure}[t]
\centering
\includegraphics[width=\linewidth]{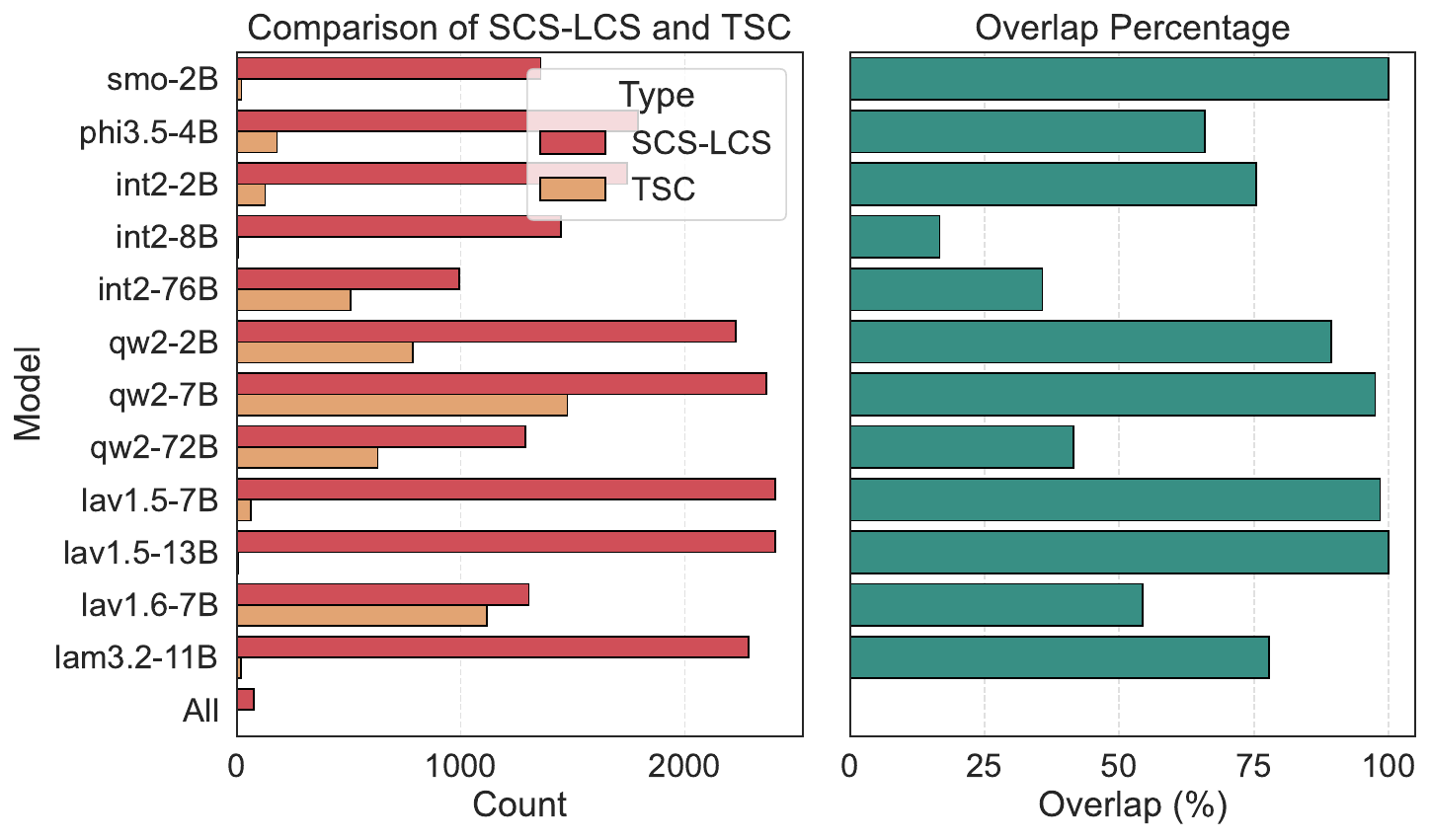}
\caption{
Overlap of neutral samples detected by the TSC and SCS-LCS methods across all models.
The left panel shows the number of neutral samples identified by each method, with TSC identifying fewer neutral samples.
The right panel presents the Min-Set Jaccard Ratio for each model's neutral sample sets, illustrating the degree of consistency between the two methods.
}
\label{fig:overlap-across-method}
\end{figure}

\subsection{Metric}
For the TSC task, we aggregate results from multiple prompt variants to identify neutral samples.
For the SCS and LCS tasks, which prompt the model to score from sarcasm and literal perspectives respectively, a sample is classified as neutral if the scores from both perspectives diverge.
We refer to this method of identifying neutral samples as SCS-LCS.
To compare the neutral samples identified by both methods, we calculate the total number of neutral samples detected by each method and assess their overlap using the Min-Set Jaccard Ratio~\footnote{The Min-Set Jaccard Ratio (MSNO) is computed as the size of the intersection divided by the size of the smaller set, offering a measure of how much the smaller set is contained within the larger set.}.
Additionally, we analyze the proportion of sarcastic and non-sarcastic ground truth labels among neutral samples identified by each method to examine which type the model is more likely to classify as neutral.

\subsection{Result}
As shown in the left panel of Figure~\ref{fig:overlap-across-method}, the number of neutral samples identified by TSC is notably low.
This suggests that directly prompting the model for neutral samples yields fewer instances, perhaps indicating the challenge of detecting ambiguity through direct classification.
The ``All'' section shows the intersection of neutral samples identified across all models.
Here, we observe that the intersection size for TSC is zero, while for the SCS-LCS method, the intersection is also quite small.
This indicates that models are largely inconsistent in identifying neutral samples, suggesting that neutral samples are often context-dependent and difficult to detect reliably across different model architectures.
\begin{figure}
\centering
\includegraphics[width=\linewidth]{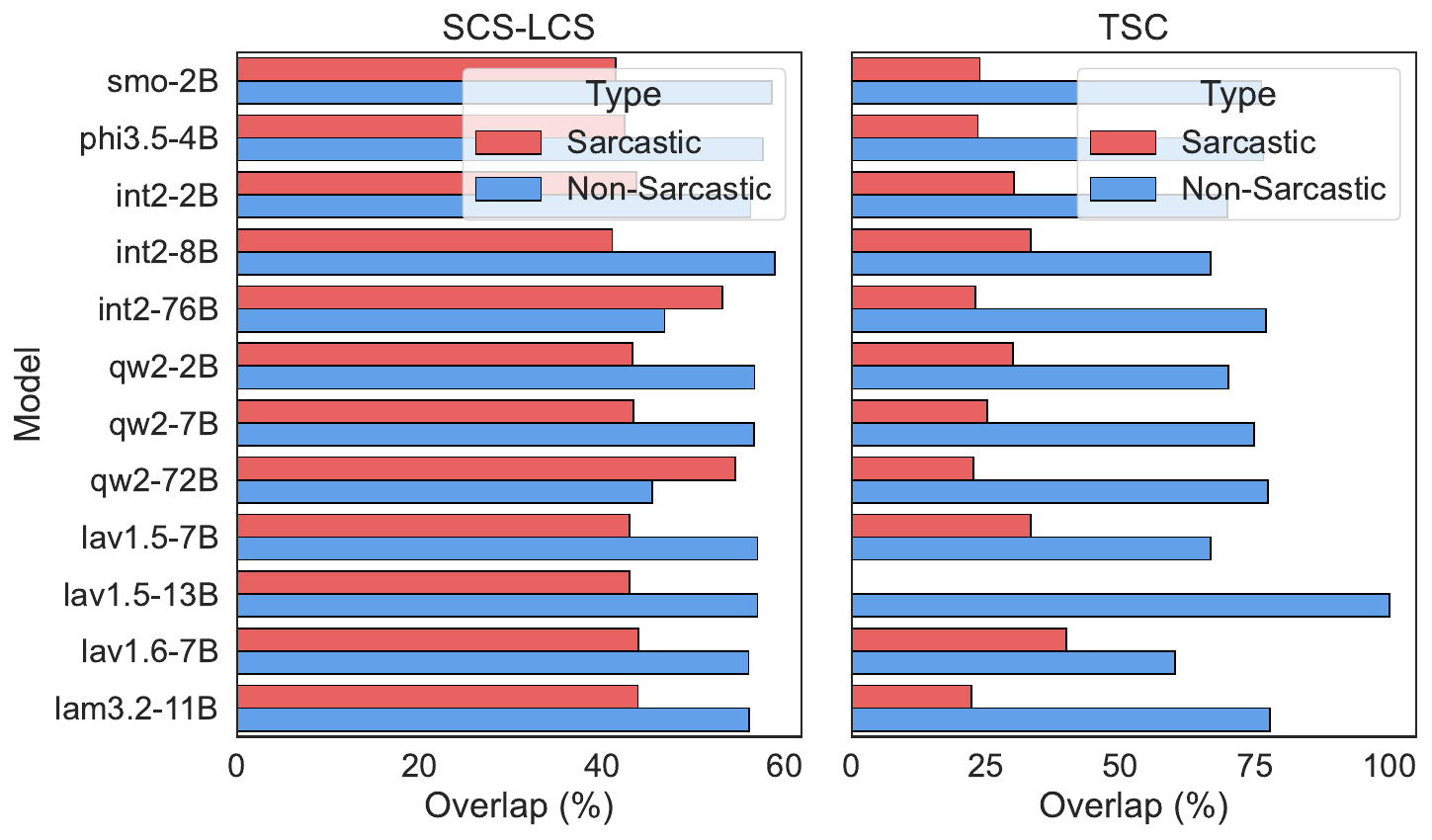}
\caption{
Proportion of ground truth labels (sarcastic vs. non-sarcastic) among the neutral samples identified by the TSC and SCS-LCS methods.
}
\label{fig:overlap-across-gt}
\end{figure}
In the right panel of Figure~\ref{fig:overlap-across-method}, we present the Jaccard similarity between the neutral sample sets identified by both methods for each model.
We find that the \texttt{qw2-7B} model exhibits relatively high and balanced consistency between the two methods, suggesting that this model is better at identifying neutral samples across different prompt strategies.
In contrast, models like \texttt{int2-8B} and \texttt{int2-76B} show significant divergence in how they identify neutral samples, implying that these models may be more sensitive to prompt variation or less stable in handling ambiguous cases.

Finally, Figure~\ref{fig:overlap-across-gt} shows that for both the SCS-LCS and TSC methods, a larger proportion of neutral samples have a non-sarcastic ground truth label.
This confirms the tendency of models to lean towards classifying more non-sarcastic samples as neutral rather than sarcastic ones.
This suggests that models might be more conservative in identifying neutral cases, potentially because sarcasm detection involves uncertainty and context dependency.

\section{Qualitative Analysis}
In this section, we present examples to demonstrate the subjectivity in multimodal sarcasm interpretation across different LVLMs and within models.
These variations highlight the challenges in establishing a definitive ground truth for sarcasm, as different models may prioritize distinct contextual cues.

\subsection{Intra-Model Subjectivity}
\begin{figure}
\centering
\includegraphics[width=\linewidth]{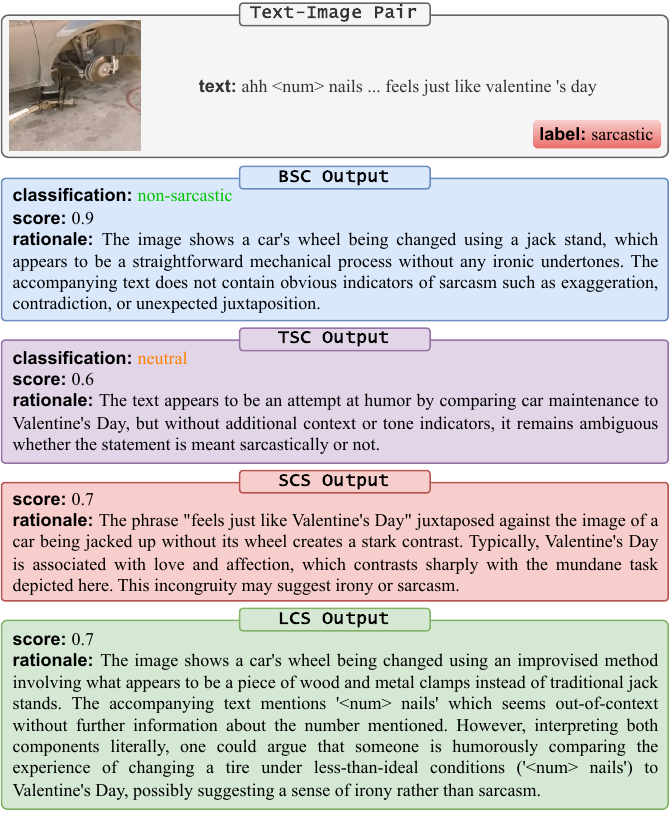}
\caption{
This image shows how the \texttt{qw2-72B} model interprets a sarcastic sample differently in the four tasks.
The model gives different classifications (non-sarcastic, neutral, or sarcastic) with different explanations.
}
\label{fig:case-all-task}
\end{figure}
Figure~\ref{fig:case-all-task} provides a concrete example where the \texttt{qw2-72B} model performs all four tasks on a sample labeled as \textit{sarcastic}.
In the BSC task, the model classifies the sample as \textit{non-sarcastic} with high confidence, suggesting that it does not detect strong irony in the given text-image pair.
The rationale indicates that the image of a car undergoing maintenance is interpreted as a straightforward mechanical process, without any clear contradiction or exaggeration in the caption.
In the TSC task, where a \textit{neutral} category is introduced, the model assigns the sample a \textit{neutral} classification.
This suggests that the text-image combination lacks definitive sarcasm cues and can be seen as ambiguous.
For the SCS and LCS tasks, the model demonstrates the ability to analyze the sample from both \textit{sarcastic} and \textit{non-sarcastic} perspectives.
Under SCS, it identifies an ironic contrast between Valentine's Day (a symbol of affection) and the mundane task of car repair, which may imply sarcasm.
Under LCS, the model takes a more literal approach, interpreting the comparison as humor rather than sarcasm.
This case underscores how task framing significantly influences model interpretation, reinforcing the idea that sarcasm is not an absolute label but a context-dependent phenomenon.
If a model can be persuaded to justify multiple perspectives depending on the prompt structure, then sarcasm detection should not be constrained to rigid binary labels.
Instead, a more flexible, spectrum-based labeling system may better capture the complexity of sarcasm in multimodal communication.

\subsection{Inter-Model Subjectivity}
\begin{figure}
\centering
\includegraphics[width=\linewidth]{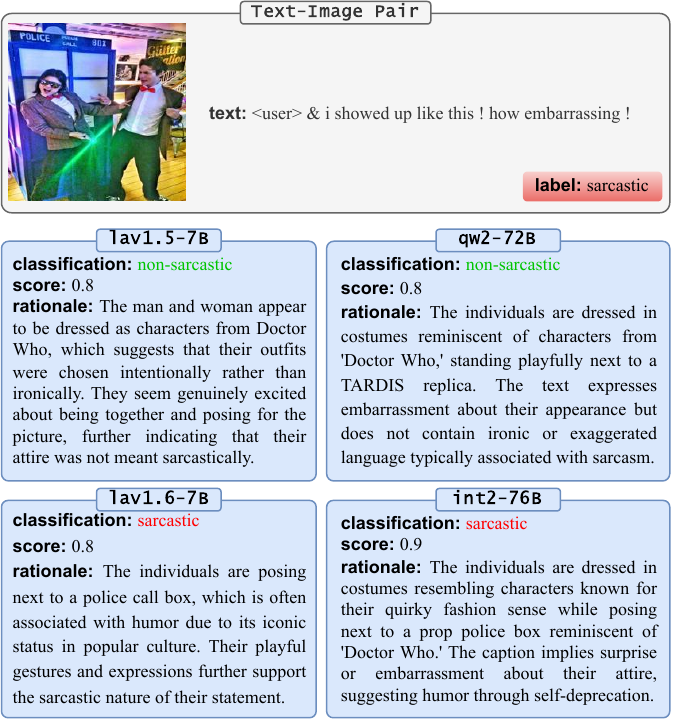}
\caption{
An example illustrating inter-model subjectivity in sarcasm interpretation.
Four LVLMs perform the BSC task on the same sarcastic text-image pair but yield divergent classifications with well-justified rationales.
While \texttt{lav1.5-7B} and \texttt{qw2-72B} interpret the scene literally, citing intentional cosplay and absence of exaggerated tone, \texttt{lav1.6-7B} and \texttt{int2-76B} emphasize playful self-deprecation and cultural irony.
This example demonstrates that different models prioritize different contextual cues (e.g., tone, attire, facial expression), supporting our claim that multimodal sarcasm is inherently subjective and that rigid binary labeling may fail to capture this diversity.
}
\label{fig:case-inter-model-subjectivity}
\end{figure}

Figure~\ref{fig:case-inter-model-subjectivity} demonstrates how different LVLMs produce varying classifications and explanations when performing the BSC task on the same sample.
Two models classified the sample as \textit{non-sarcastic}, while the other two classified it as \textit{sarcastic}.
Despite these opposing classifications, the rationales provided by all models appear reasonable.
This suggests that sarcasm detection is inherently subjective, as different models may prioritize different contextual cues when making their predictions.
Furthermore, the dataset's ground truth labeling does not account for the fact that some samples can be interpreted from both sarcastic and non-sarcastic perspectives.
This raises concerns about the binary labeling approach, which may oversimplify the nuanced nature of sarcasm perception in multimodal data.
A rigid binary classification scheme overlooks the possibility that sarcasm exists on a spectrum, where ambiguity is often a defining characteristic.

\section{A Mini Mixed Dataset for Enhanced Evaluation}

To complement our main evaluation and address key concerns regarding dataset diversity, prompt robustness, and model reliability, we conduct a focused small-scale experiment designed to test the generalizability and interpretive consistency of our framework under extended conditions.
Specifically, we aim to investigate whether our core findings hold when applied to additional datasets with different stylistic and contextual features, a broadened set of prompt variants for each task, and representative commercial vision-language models not included in the main study.

Due to computational constraints and commercial API limitations, we restrict this supplementary analysis to a 100-sample mini-benchmark.
Despite its limited scale, this targeted setup enables us to isolate and examine critical dimensions of model behavior, including prompt sensitivity, cross-dataset generalization, and the alignment of model outputs with human judgment, with improved granularity and interpretability.

\subsection{Dataset Construction}\label{sec:mini-dataset}

To assess the generalizability of our framework beyond a single benchmark, we construct a small-scale evaluation set by sampling from two distinct multimodal sarcasm datasets: MMSD2.0~\cite{qin2023mmsd2} and DocMSU~\cite{Du2023DocMSUAC}.

MMSD2.0 is a widely used benchmark for multimodal sarcasm detection, featuring carefully curated image-text pairs with binary sarcastic and non-sarcastic labels. Compared to its predecessor MMSD~\cite{cai2019multi}, MMSD2.0 mitigates spurious textual cues and label noise, making it a more reliable dataset for sarcasm analysis.
DocMSU is a document-level benchmark designed to evaluate multimodal sarcasm understanding in longer and more context-rich samples. It includes sarcastic and non-sarcastic posts from Reddit~\footnote{\url{https://www.reddit.com}}, combining textual content with associated images or memes, and reflects more diverse linguistic and visual patterns.
From each dataset, we randomly sample 25 sarcastic and 25 non-sarcastic instances, resulting in a balanced 100-sample mini-benchmark. This combined set captures stylistic variation across platforms, annotation strategies, and sarcasm manifestations, and provides a compact yet diverse basis for extended evaluation.

\subsection{Prompt Variant Scaling}

While our main study uses three semantically equivalent prompt variants per task to evaluate inter-prompt consistency, some readers may reasonably question whether this number is sufficient to reveal the full extent of prompt sensitivity in LVLMs.
After all, surface-level phrasing can introduce subtle but consequential variation in how models interpret the same task, particularly in subjective domains such as sarcasm~\cite{sclar2024quantifying, mizrahi2024state, zhang-etal-2025-prompt-design}.
To further explore this dimension, we expand each task to include ten prompt variants, all designed to be semantically equivalent but lexically diverse.
This setup allows for a more rigorous examination of consistency under increased linguistic variation and provides a broader empirical basis for evaluating model robustness.
We focus this experiment on two representative model families: Qwen2-VL~\cite{Qwen2VL} and InternVL2~\cite{chen2024far}.
These families were selected due to their strong performance in earlier experiments and their architectural contrast.
Qwen2 models consistently demonstrated stable classification and rationale behavior, while InternVL2 models showed higher variability, making them ideal for studying prompt-induced instability.

\begin{figure}
\centering
\includegraphics[width=0.9\linewidth]{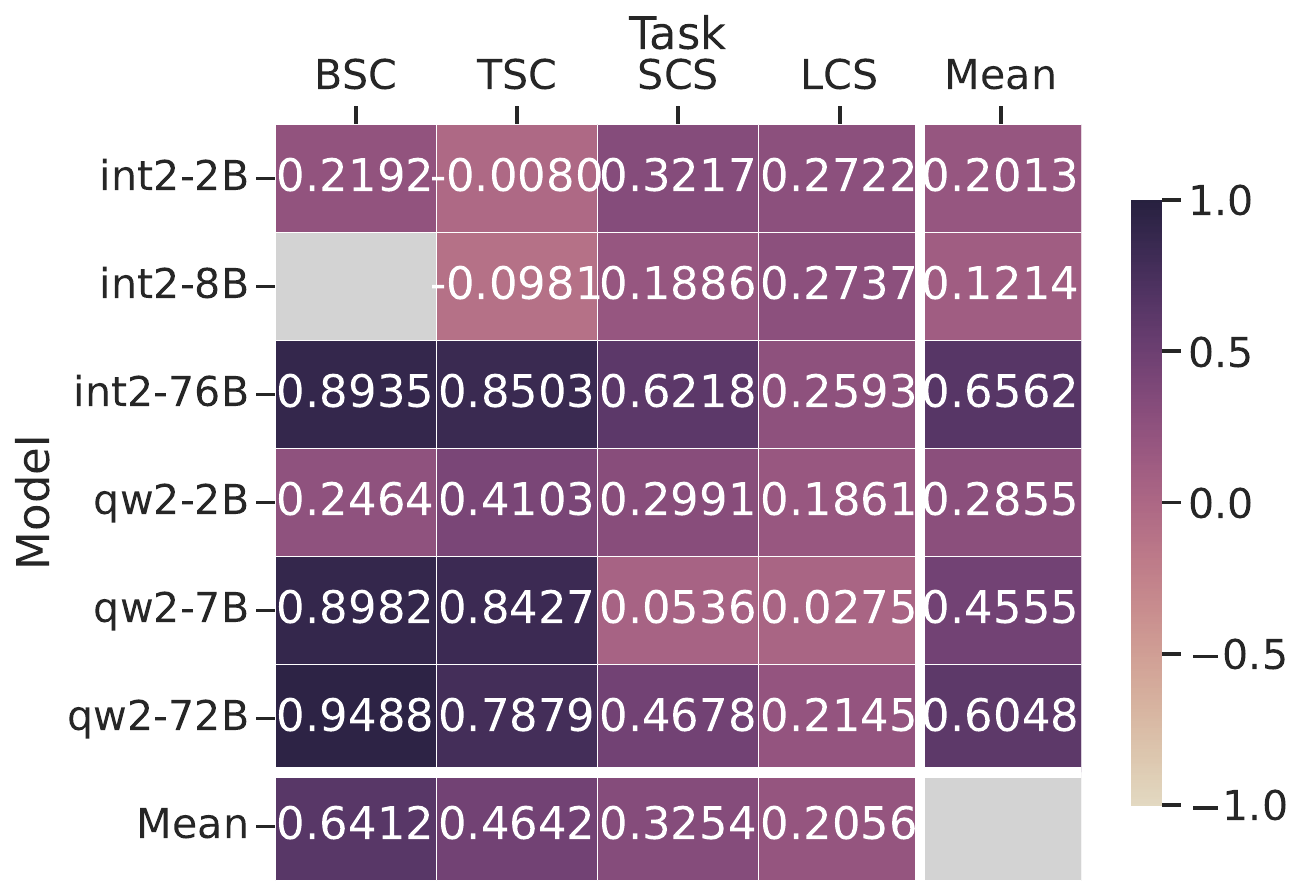}
\caption{
Classification consistency heatmap under expanded prompt settings.
}
\label{fig:cls-consistency-mini-10}
\end{figure}

\begin{figure}
\centering
\includegraphics[width=0.72\linewidth]{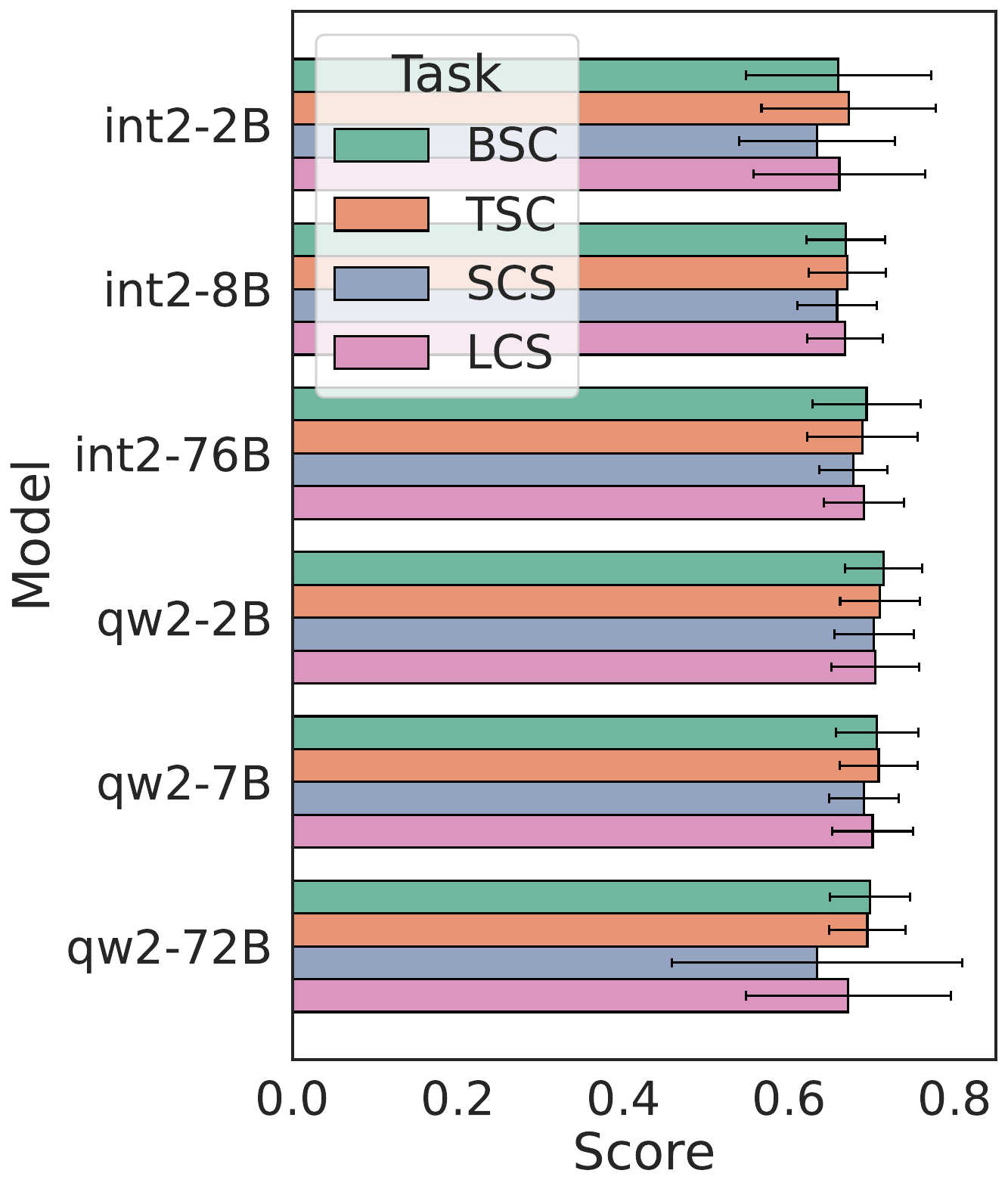}
\caption{
Rationale consistency scores under extended prompt diversity.
}
\label{fig:rationale-consistency-mini-10}
\end{figure}

As shown in Figure~\ref{fig:cls-consistency-mini-10}, we observe a general decline in classification consistency scores across tasks compared to the 3-prompt setting.
Even in tasks like BSC and TSC, which are structurally more constrained, consistency drops for several models (e.g., \texttt{int2-76B}, \texttt{qw2-7B}).
This suggests that prompt sensitivity is more substantial than previously estimated, highlighting that limited prompt variants may obscure important variation in model behavior.
Similarly, rationale consistency (Figure~\ref{fig:rationale-consistency-mini-10}) declines slightly across models and tasks, especially in SCS and LCS. Despite this, Qwen2 models retain a higher level of rationale stability than InternVL counterparts, echoing previous observations that model architecture plays a pivotal role in interpretive robustness.
These findings confirm that prompt phrasing has non-trivial effects on sarcasm interpretation in LVLMs, and that increasing prompt diversity exposes more variation than smaller prompt sets reveal.
We argue that future evaluations, especially for tasks involving subjectivity, should systematically account for prompt variation as a core evaluation axis rather than treating it as a minor implementation detail.

\subsection{Evaluation on Commercial LVLMs}

\begin{figure*}[t]
\centering
\includegraphics[width=1\linewidth]{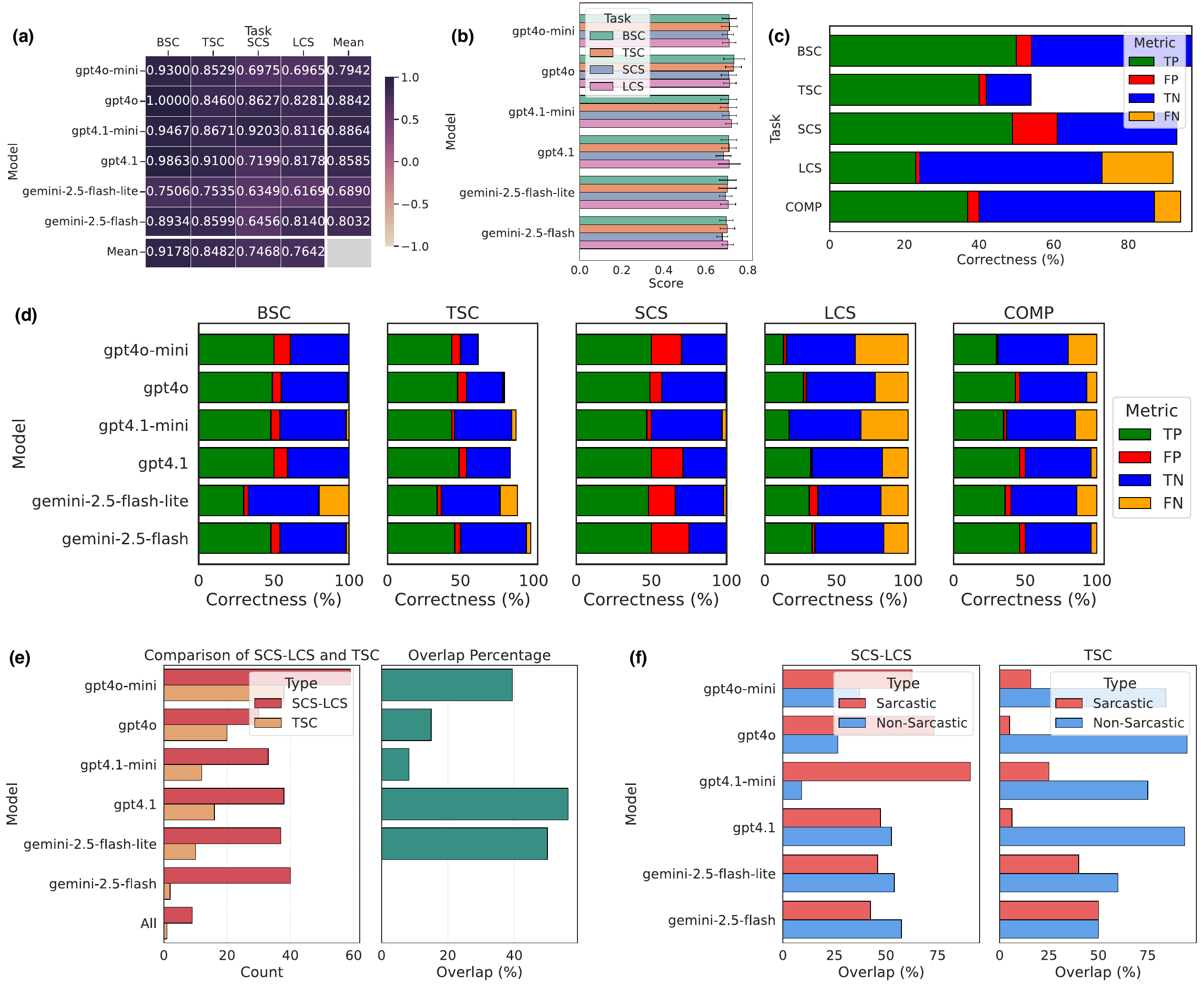}
\caption{
Quantitative analysis of commercial models.
}
\label{fig:commercial-eval}
\end{figure*}

To complement our analysis of open-source LVLMs and provide a more comprehensive picture of current multimodal sarcasm capabilities, we evaluate leading commercial models that represent the state-of-the-art in instruction-following and visual reasoning.
While our primary experiments focus on publicly available models for reproducibility, commercial systems often demonstrate superior alignment and reasoning abilities, making them valuable baselines for this task.
Specifically, we test six proprietary models: \textbf{GPT-4o}, \textbf{GPT-4o-mini}, \textbf{GPT-4.1}, \textbf{GPT-4.1-mini}, \textbf{Gemini-2.5-Flash}, and \textbf{Gemini-2.5-Flash-lite}.
These models span different latency-accuracy trade-offs and are deployed via official APIs.
All models are evaluated using the original three prompt variants per task on the 100-sample mini-benchmark introduced in Section~\ref{sec:mini-dataset}.
This setup allows us to directly compare their behavior to open-source models under a consistent protocol.

We first analyze the inter-prompt consistency of commercial LVLMs across the four evaluation tasks.
As shown in Figure~\ref{fig:commercial-eval}(a) and (b), these models exhibit notably high agreement across semantically equivalent prompts.
Krippendorff’s $\alpha$ scores remain positive and robust for all models, with near-perfect consistency observed in the BSC and TSC tasks for GPT-4o and GPT-4.1.
This suggests that commercial systems possess strong surface-level prompt invariance, likely supported by more extensive instruction tuning and alignment strategies.
However, despite this overall stability, a consistent drop is observed in the SCS and LCS tasks, where consistency is lower than in BSC and TSC across all models.
This mirrors patterns seen in open-source models and points to a broader phenomenon: models, regardless of size or training, struggle more with interpretive tasks that require reasoning from a constrained perspective (e.g., sarcasm-only or literal-only), where subjective ambiguity cannot be resolved purely through prompt structure.
One plausible explanation is that while commercial models are better aligned for general-purpose instruction following, they may default to a balanced interpretation strategy when faced with highly contextual or multi-intent stimuli, which undermines strict perspective adherence under prompt variation.
Rationale consistency follows a similar trend: although overall coherence remains high, the gap between BSC/TSC and SCS/LCS again indicates that open-ended interpretive generation is inherently more sensitive to prompt phrasing, even for high-end models.
These findings suggest that while alignment improves robustness to surface phrasing, interpretive rigidity under subjective constraints remains a fundamental challenge.

We then examine how commercial LVLMs compare with ground-truth labels, not as an absolute measure of performance, but to observe systematic patterns in model predictions.
As shown in Figure~\ref{fig:commercial-eval}(c) and (d), BSC task results are largely consistent with the annotated labels, with most errors arising from false positives, suggesting a slight tendency to over-assign sarcasm.
In TSC, correctness decreases, especially for GPT models, mainly due to neutral classifications being excluded from evaluation, which highlights a mismatch between model uncertainty and rigid binary labels.
SCS and LCS reveal distinct interpretive biases: sarcasm-centric prompting yields more false positives, while literal-centric prompting leads to sarcasm underprediction through false negatives.
These trends persist in the COMP setting, where all models lean toward literal readings.
These observations align with those found in open-source models, suggesting that even advanced commercial systems are shaped by the same label ambiguity and framing asymmetries inherent in sarcasm detection.

Finally, we analyze how commercial models identify neutral samples using the TSC and SCS-LCS strategies.
As shown in Figure~\ref{fig:commercial-eval}(e), commercial models detect relatively few neutral samples overall, with TSC consistently identifying fewer instances than SCS-LCS.
This mirrors the behavior observed in open-source models and underscores the difficulty of eliciting ambiguity through direct classification.
Among the models, GPT-4.1 and Gemini-2.5 Flash exhibit the highest consistency between the two methods, suggesting more stable neutral identification under different prompting conditions.
However, for most models, the Jaccard overlap remains low, indicating that what is perceived as ``neutral'' still varies substantially across strategies even in aligned systems.
Figure~\ref{fig:commercial-eval}(f) further reveals that, similar to open-source models, neutral predictions in both methods are disproportionately drawn from non-sarcastic ground-truth samples.
This bias suggests a general conservativeness in commercial models: when uncertain, they tend to interpret ambiguity as a weakened form of literalness rather than potential sarcasm.
While this may reflect a rational default under annotation uncertainty, it also reinforces the limitations of using binary-labeled datasets for evaluating subjective phenomena like sarcasm, where ambiguity may legitimately arise on both sides of the label boundary.

In summary, commercial LVLMs show higher consistency and more stable behavior than open-source models, but still exhibit similar interpretive biases and difficulty with subjective prompts. We omit confidence-based analyses (e.g., NLL) in this section, as several commercial APIs do not provide log-probabilities, making direct comparison infeasible.

\subsection{Human Evaluation}

\begin{figure*}[t]
\centering
\includegraphics[width=\linewidth]{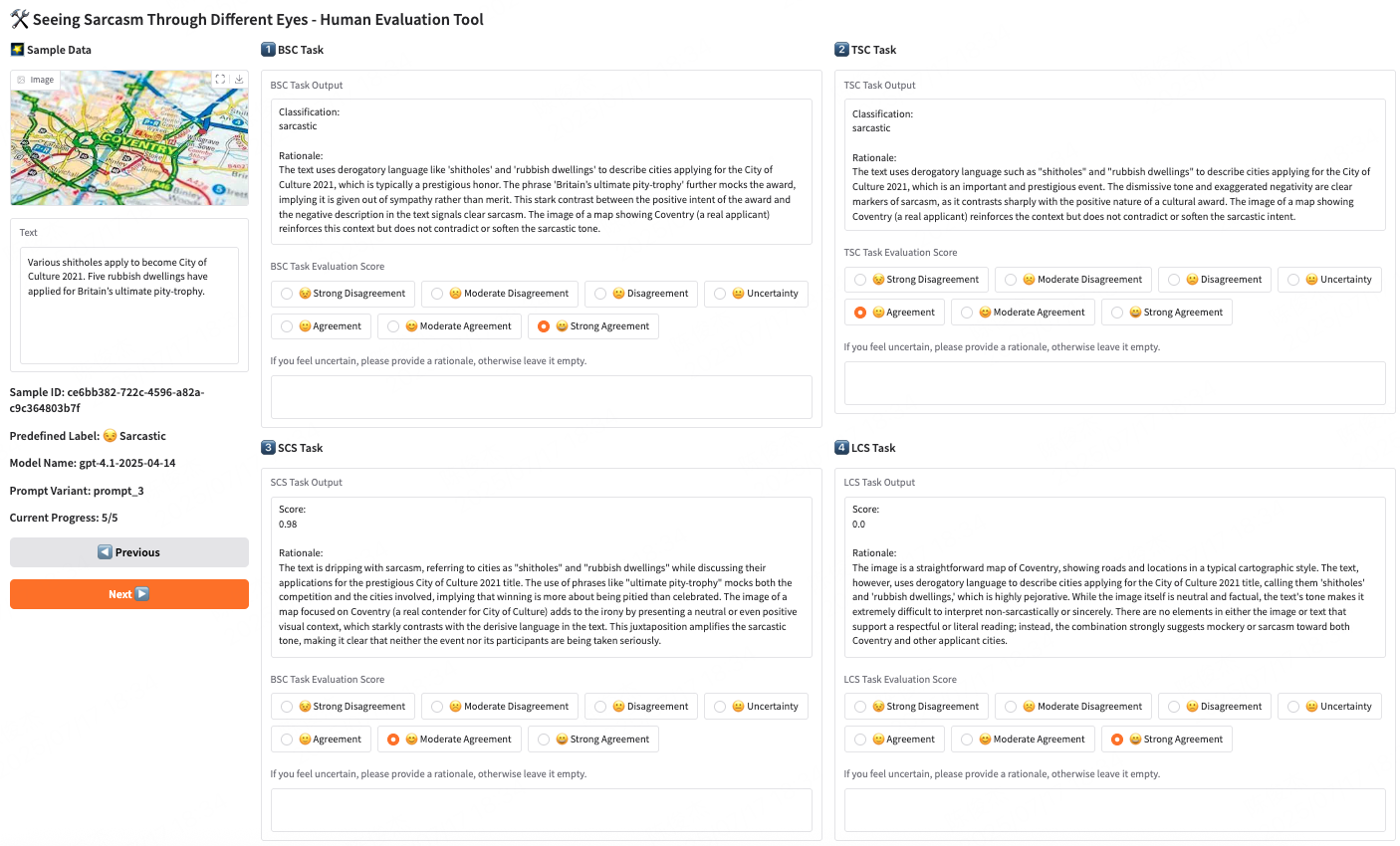}
\caption{UI interface of the human evaluation tool.}
\label{fig:human-eval-tool}
\end{figure*}

\begin{table}[t]
\centering
\caption{Distribution of Human Ratings Across Models and Tasks (\%).}
\begin{tabularx}{\linewidth}{l|XXXX}
\toprule
\textbf{Rating} & \textbf{BSC} & \textbf{TSC} & \textbf{SCS} & \textbf{LCS} \\
\midrule
\multicolumn{5}{l}{\textbf{GPT-4o}} \\
Strong Disagr. (-3) & \cellcolor[RGB]{247,251,255}0.00 & \cellcolor[RGB]{247,251,255}0.00 & \cellcolor[RGB]{247,251,255}0.00 & \cellcolor[RGB]{246,250,254}0.33 \\
Mod. Disagr. (-2) & \cellcolor[RGB]{247,251,255}0.00 & \cellcolor[RGB]{247,251,255}0.11 & \cellcolor[RGB]{247,251,255}0.00 & \cellcolor[RGB]{244,249,254}0.67 \\
Disagreement (-1) & \cellcolor[RGB]{243,248,253}0.78 & \cellcolor[RGB]{234,242,250}3.22 & \cellcolor[RGB]{237,244,251}2.33 & \cellcolor[RGB]{227,238,248}4.89 \\
Uncertainty (0) & \cellcolor[RGB]{247,251,255}0.00 & \cellcolor[RGB]{247,251,255}0.00 & \cellcolor[RGB]{247,251,255}0.00 & \cellcolor[RGB]{246,250,254}0.22 \\
Agreement (+1) & \cellcolor[RGB]{85,159,205}27.67 & \cellcolor[RGB]{39,119,184}35.44 & \cellcolor[RGB]{57,137,193}32.22 & \cellcolor[RGB]{42,122,185}35.00 \\
Mod. Agreement (+2) & \cellcolor[RGB]{8,48,107}49.00 & \cellcolor[RGB]{26,105,174}38.22 & \cellcolor[RGB]{8,73,145}44.11 & \cellcolor[RGB]{47,127,188}33.89 \\
Strong Agreement (+3) & \cellcolor[RGB]{123,183,217}22.56 & \cellcolor[RGB]{119,180,216}23.00 & \cellcolor[RGB]{133,188,219}21.33 & \cellcolor[RGB]{103,171,212}25.00 \\
\midrule
\multicolumn{5}{l}{\textbf{GPT-4.1}} \\
Strong Disagr. (-3) & \cellcolor[RGB]{247,251,255}0.00 & \cellcolor[RGB]{247,251,255}0.11 & \cellcolor[RGB]{247,251,255}0.11 & \cellcolor[RGB]{247,251,255}0.11 \\
Mod. Disagr. (-2) & \cellcolor[RGB]{247,251,255}0.11 & \cellcolor[RGB]{247,251,255}0.11 & \cellcolor[RGB]{246,250,254}0.33 & \cellcolor[RGB]{243,248,253}0.78 \\
Disagreement (-1) & \cellcolor[RGB]{243,248,253}1.00 & \cellcolor[RGB]{241,247,253}1.44 & \cellcolor[RGB]{230,240,249}4.11 & \cellcolor[RGB]{225,237,248}5.22 \\
Uncertainty (0) & \cellcolor[RGB]{247,251,255}0.00 & \cellcolor[RGB]{247,251,255}0.00 & \cellcolor[RGB]{247,251,255}0.00 & \cellcolor[RGB]{247,251,255}0.11 \\
Agreement (+1) & \cellcolor[RGB]{74,151,201}29.33 & \cellcolor[RGB]{55,135,192}32.67 & \cellcolor[RGB]{46,126,188}34.11 & \cellcolor[RGB]{49,129,189}33.67 \\
Mod. Agreement (+2) & \cellcolor[RGB]{8,57,120}47.11 & \cellcolor[RGB]{22,99,170}39.33 & \cellcolor[RGB]{30,109,178}37.44 & \cellcolor[RGB]{42,122,185}34.89 \\
Strong Agreement (+3) & \cellcolor[RGB]{123,183,217}22.44 & \cellcolor[RGB]{94,165,209}26.33 & \cellcolor[RGB]{112,177,215}23.89 & \cellcolor[RGB]{102,170,212}25.22 \\
\midrule
\multicolumn{5}{l}{\textbf{Gemini-2.5-Flash}} \\
Strong Disagr. (-3) & \cellcolor[RGB]{247,251,255}0.00 & \cellcolor[RGB]{247,251,255}0.00 & \cellcolor[RGB]{246,250,254}0.22 & \cellcolor[RGB]{247,251,255}0.00 \\
Mod. Disagr. (-2) & \cellcolor[RGB]{247,251,255}0.11 & \cellcolor[RGB]{247,251,255}0.00 & \cellcolor[RGB]{245,249,254}0.56 & \cellcolor[RGB]{245,249,254}0.44 \\
Disagreement (-1) & \cellcolor[RGB]{243,248,253}0.89 & \cellcolor[RGB]{243,248,253}0.78 & \cellcolor[RGB]{217,231,245}7.33 & \cellcolor[RGB]{233,242,250}3.33 \\
Uncertainty (0) & \cellcolor[RGB]{247,251,255}0.11 & \cellcolor[RGB]{247,251,255}0.11 & \cellcolor[RGB]{247,251,255}0.00 & \cellcolor[RGB]{247,251,255}0.11 \\
Agreement (+1) & \cellcolor[RGB]{79,155,203}28.67 & \cellcolor[RGB]{52,132,191}33.00 & \cellcolor[RGB]{61,141,195}31.44 & \cellcolor[RGB]{75,152,201}29.22 \\
Mod. Agreement (+2) & \cellcolor[RGB]{8,62,128}46.22 & \cellcolor[RGB]{12,86,160}41.78 & \cellcolor[RGB]{32,112,180}36.78 & \cellcolor[RGB]{28,107,176}37.78 \\
Strong Agreement (+3) & \cellcolor[RGB]{111,176,214}24.00 & \cellcolor[RGB]{107,174,214}24.33 & \cellcolor[RGB]{114,177,215}23.67 & \cellcolor[RGB]{75,152,201}29.11 \\
\bottomrule
\end{tabularx}
\label{tab:human-eval-colored}
\end{table}

\begin{table}[t]
\centering
\caption{Inter-Annotator Agreement (Krippendorff's $\alpha$) across Tasks and Models.}
\begin{threeparttable}
\begin{tabularx}{\linewidth}{l|XXXX}
\toprule
\textbf{Model} & \textbf{BSC} & \textbf{TSC} & \textbf{SCS} & \textbf{LCS} \\
\midrule
GPT-4o & \cellcolor[RGB]{106,173,213}0.14 & \cellcolor[RGB]{187,214,235}0.07 & \cellcolor[RGB]{187,214,235}0.07 & \cellcolor[RGB]{106,173,213}0.14 \\
GPT-4.1 & \cellcolor[RGB]{57,137,193}0.19 & \cellcolor[RGB]{247,251,255}-0.02 & \cellcolor[RGB]{41,121,185}0.21 & \cellcolor[RGB]{8,48,107}0.30 \\
Gemini-2.5-Flash & \cellcolor[RGB]{8,55,117}0.29 & \cellcolor[RGB]{240,246,252}-0.01 & \cellcolor[RGB]{48,128,189}0.20 & \cellcolor[RGB]{203,222,240}0.05 \\
\bottomrule
\end{tabularx}
\label{tab:interannotator-agreement}
\begin{tablenotes}
\footnotesize
\item[*] Inter-annotator agreement (Krippendorff’s $\alpha$) is computed based on a three-level categorization of human ratings: agreement (scores +1, +2, +3), uncertainty (score 0), and disagreement (scores –1, –2, –3). This abstraction reduces Likert-scale granularity to focus on high-level consensus.
\end{tablenotes}
\end{threeparttable}
\end{table}

Given the inherently subjective nature of sarcasm, we complement our automated evaluation with a human assessment protocol to examine whether model outputs are subjectively plausible. Instead of asking annotators to perform the tasks themselves, we focus on validating the reasonableness of the model-generated predictions and rationales from a human perspective.
To support this process, we design a custom human evaluation tool that allows annotators to view each multimodal input along with the model's classification, score, and rationale for all four tasks.
As shown in Figure~\ref{fig:human-eval-tool}, the interface presents the image, accompanying text, model metadata, and task-specific outputs, followed by a 7-point Likert-style agreement scale ranging from \textit{Strong Disagreement} to \textit{Strong Agreement}.
Annotators evaluate each task independently, judging whether the given rationale sufficiently supports the predicted label. This setup enables us to assess interpretive plausibility without requiring full task replication by human raters.

To quantify human evaluation outcomes, we report two metrics for each model and task.
First, we compute the distribution of Likert ratings to capture how plausible the model outputs appear to human judges.
This provides insight into the subjective quality and perceived reliability of model reasoning under different prompt formulations.
Second, we measure inter-annotator agreement using Krippendorff’s $\alpha$, which accommodates ordinal scales and handles missing or imbalanced data more robustly than alternative measures.
This allows us to examine whether certain tasks or models produce more ambiguous or contentious predictions.

We apply this evaluation to the outputs of three representative commercial LVLMs: GPT-4o, GPT-4.1, and Gemini-2.5-Flash.
For each model, we evaluate three prompt variants per task over the full 100-sample mini-benchmark introduced in Section~\ref{sec:mini-dataset}.
Three annotators independently assess each output, with each annotator completing 3,600 judgments (3 models × 4 tasks × 3 prompt variants × 100 samples).
This results in a total of 10,800 annotations across all annotators.

Our results reveal that the vast majority of model outputs were judged as subjectively plausible by human annotators. As shown in Table~\ref{tab:human-eval-colored}, agreement ratings of 1 or higher (i.e., \textit{Agreement}, \textit{Moderate Agreement}, \textit{Strong Agreement}) dominate across all models and tasks, particularly in the BSC setting. This indicates that, even without task-specific fine-tuning, commercial LVLMs can produce explanations that human raters find sensible and well-grounded.

From the perspective of inter-rater reliability, Table~\ref{tab:interannotator-agreement} shows that annotator agreement was consistently highest on the BSC task, with Krippendorff’s $\alpha$ scores of 0.14 (GPT-4o), 0.19 (GPT-4.1), and 0.29 (Gemini-2.5-Flash).
This confirms that straightforward sarcasm classification is the most reliably judged by humans.
In contrast, TSC yielded the lowest overall agreement, with values even falling into the negative range for GPT-4.1 and Gemini.
This reflects the challenge of interpreting neutrality consistently.
SCS and LCS also show distinctive trends across models.
In SCS, GPT-4.1 achieved relatively high inter-annotator agreement ($\alpha = 0.21$), while GPT-4o’s $\alpha = 0.07$ was the lowest, despite both models receiving largely positive human ratings.
This suggests that although GPT-4o’s rationales were generally perceived as agreeable (Table~\ref{tab:human-eval-colored}), their framing may have been more variable or ambiguous, leading to less consistent interpretation across annotators.
For LCS, Gemini-2.5-Flash showed weaker agreement ($\alpha = 0.05$), despite strong Likert scores.
This implies that annotators may disagree not about the quality of the rationale itself, but about whether it truly embodies a literal perspective.

Together, these results suggest that commercial LVLMs generate outputs broadly acceptable to humans, especially for direct classification tasks.
However, constrained interpretive tasks like SCS and LCS introduce greater subjectivity and variability, even when output plausibility remains high.
This underscores the importance of multi-perspective evaluation for understanding model behavior beyond label alignment.

\section{Discussion}
Our study demonstrates that large vision-language models (LVLMs), when evaluated through a carefully designed multi-perspective framework, reveal substantial interpretive variability in multimodal sarcasm understanding.
These variations occur not only across model architectures, but also within the same model under different prompt phrasings and task formulations.
This reinforces our central argument: sarcasm is inherently subjective and context-sensitive, and current binary-labeled datasets do not fully capture this complexity.

Through extensive experiments across 12 open-source models and a supplementary benchmark involving commercial systems and human evaluators, we identify two consistent findings:
(1) LVLMs display stronger confidence and agreement on literal interpretations than sarcastic ones, and
(2) interpretive prompts (e.g., SCS and LCS) introduce greater response variability than classification prompts (e.g., BSC and TSC), even in high-performing models.
These findings point to a need for both uncertainty-aware modeling and more flexible annotation schemes that acknowledge interpretive ambiguity.

At the same time, we recognize that our current framework adopts a strictly zero-shot prompting strategy to ensure uniform evaluation across models.
While this allows us to isolate intrinsic model behavior, it may also limit the expressive reasoning capabilities of LVLMs, particularly in tasks requiring constrained interpretive perspectives.
Prior work such as MemeGuard~\cite{jha2024memeguard} and VIGNETTE~\cite{raj2025vignette} has shown that prompting strategies like few-shot demonstrations, question decomposition, or chain-of-thought reasoning can improve model reliability in subjective or socially grounded tasks.
Incorporating such techniques into our framework could mitigate prompt sensitivity, enhance rationale consistency, and better reflect human-like reasoning.
We view these methods as promising next steps.
For example, few-shot prompting may help models internalize contextual contrast more effectively, while chain-of-thought prompts may encourage models to explicitly differentiate between literal surface meaning and ironic intent.
Additionally, socially informed prompting strategies, as proposed in VIGNETTE, could offer templates for disambiguating sarcasm from humor or criticism through guided sub-questioning.
Due to resource constraints and our focus on establishing a zero-shot benchmark, we leave the empirical validation of these methods to future work, but emphasize their conceptual alignment with the goals of our framework.

Overall, our findings provide a strong baseline for evaluating subjective multimodal phenomena with LVLMs.
They also lay the groundwork for richer prompting strategies and more cognitively aligned modeling approaches that go beyond label agreement toward interpretive plausibility and human-consistent understanding.

\section{Conclusion} 
This study introduced a four-task evaluation framework to assess how Large Vision-Language Models (LVLMs) interpret multimodal sarcasm from multiple perspectives.
Evaluating 12 open-source LVLMs on MMSD2.0, we found that models vary widely in their sarcasm judgments, with larger models showing more stability but still exhibiting interpretive bias.  
The TSC task revealed that many samples fall between sarcastic and non-sarcastic, highlighting the limitations of binary labels.
Models also tended to favor literal interpretations unless explicitly guided otherwise.
To test generalizability, we conducted a supplementary experiment on a 100-sample mini-benchmark combining MMSD2.0 and DocMSU, using more diverse prompts and including commercial LVLMs like GPT-4o.
The results confirmed our main findings, and a human evaluation further showed that most model outputs were perceived as plausible, though variability increased under interpretive constraints.
Overall, our findings call for sarcasm evaluation frameworks that move beyond binary classification toward uncertainty-aware, multi-perspective modeling that better reflects human interpretive diversity.

\section*{Acknowledgments}
The research presented in this paper was funded by the Anhui Provincial Natural Science Foundation under Grant No. 2308085MF220.
It also received additional support from the Anhui University Natural Science Foundation through Grant No. 2023AH050914.

\appendices

\section{Detailed Results}\label{app:agreement-with-ground-truth-results}
Due to space constraints, we do not include detailed tabular representations of the raw data underlying each figure in the main text. Instead, to ensure full transparency and reproducibility, all such data are made publicly available in our open-source repository\footnote{\url{https://github.com/CoderChen01/LVLMSarcasmAnalysis/tree/main/lvlm_sarcasm_analysis/raw_data}}.

\section{Prompt Details}\label{app:prompt-details}
Due to space limitations, we refer readers to our open-source repository\footnote{\url{https://github.com/CoderChen01/LVLMSarcasmAnalysis/tree/main/lvlm_sarcasm_analysis/prompts}}
for all prompt variants.
For each task-specific prompt, we adopt a structured approach to ensure clarity and consistency in model evaluation. First, we provide a concise description of the task, clearly defining the objective and scope. Next, we outline the analytical steps required for the model to perform the task effectively, guiding its reasoning process in a more structured manner. Finally, we specify the expected output format, ensuring that the model's responses adhere to a standardized structure, making it easier to compare results across different tasks and models.

\bibliographystyle{IEEEtran}
\bibliography{reference}



\newpage
\begin{IEEEbiography}[{\includegraphics[width=1in,height=1.25in,clip,keepaspectratio]{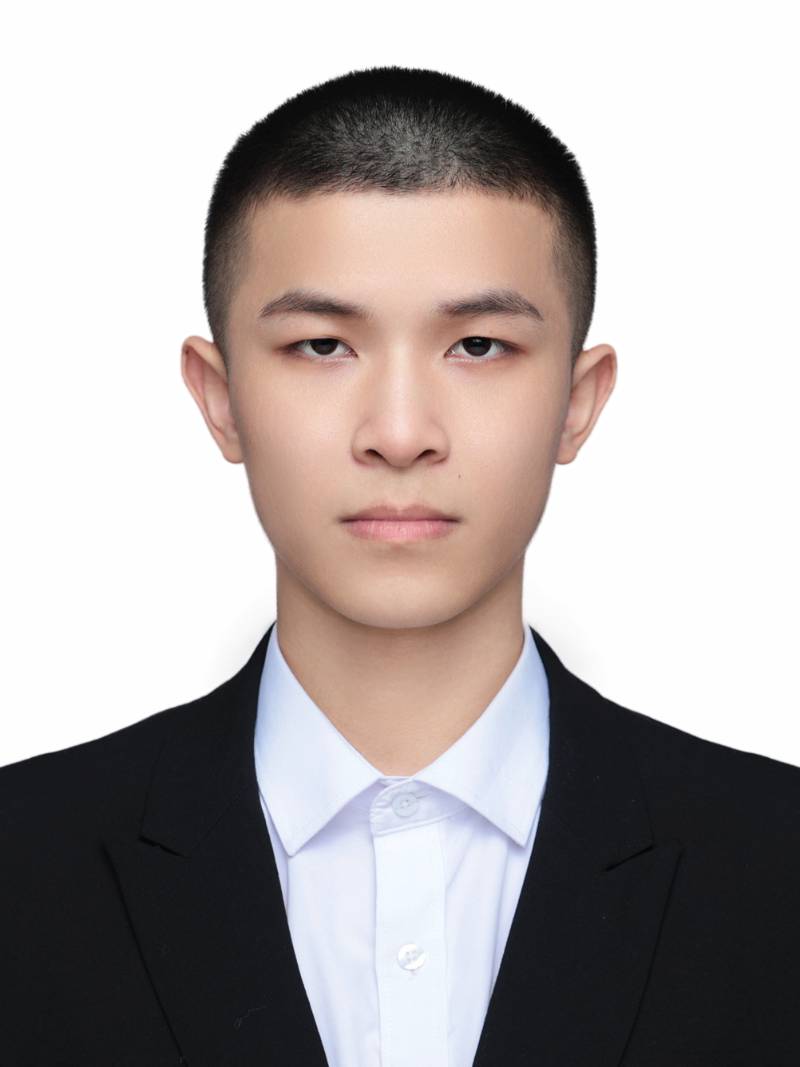}}]{Junjie Chen} received the B.S. degree at Anhui Polytechnic University~(AHPU), Wuhu, China, in 2023.
He is currently pursuing the M.S. degree at AHPU, under the supervision of Associate Professor Subin Huang, and is expected to graduate in 2026.

His research interests include efficient deep learning, multi-modal deep learning, and their applications in artificial intelligence.
\end{IEEEbiography}


\begin{IEEEbiography}[{\includegraphics[width=1in,height=1.25in,clip,keepaspectratio]{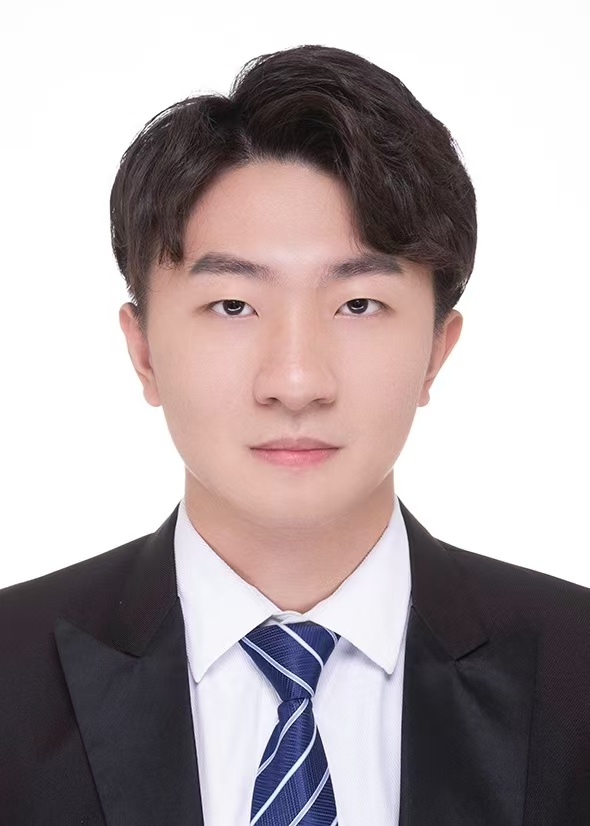}}]{Xuyang Liu} received the B.S. degree in transportation engineering from Hebei University of Technology, Tianjin, China, in 2023. He is currently pursuing the M.S. degree with the College of Electronics and Information Engineering, Sichuan University, Chengdu, China, under the supervision of Prof. Honggang Chen.

His current research interests include vision-language models, transfer learning, and model compression.
\end{IEEEbiography}


\begin{IEEEbiography}[{\includegraphics[width=1in,height=1.25in,clip,keepaspectratio]{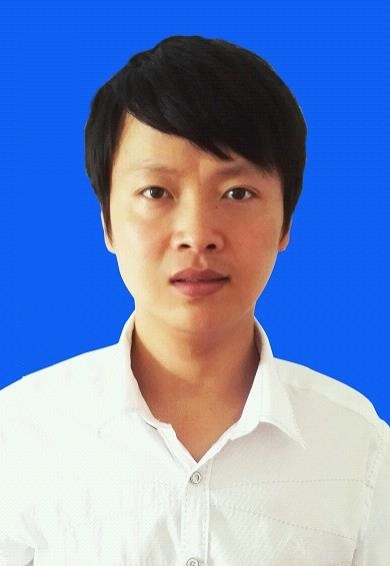}}]{Subin Huang} received his Ph.D. degree from the School of Computer Engineering and Science at Shanghai University, China, in 2020.

He is an Associate Professor at the School of Computer and Information at Anhui Polytechnic University in China.

His research primarily focuses on information retrieval, data mining, and multi-modal learning.
\end{IEEEbiography}


\begin{IEEEbiography}[{\includegraphics[width=1in,height=1.25in,clip,keepaspectratio]{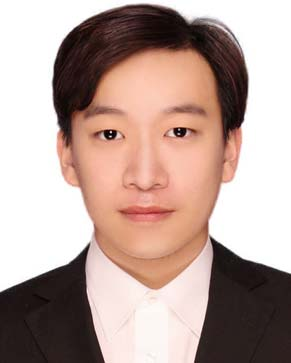}}]{Linfeng Zhang}
received the Ph.D. degree in June 2024 from the Institute for Interdisciplinary Information Sciences (IIIS), Tsinghua University, Beijing, China, where he was supervised by Prof. Kaisheng Ma. 
He has published papers in top-tier conferences such as NeurIPS, ICLR, CVPR, and ICCV, as well as in journals like IEEE TPAMI.

He is an Assistant Professor at the School of Artificial Intelligence, Shanghai Jiao Tong University, Shanghai, China.

His research interests include computer vision, deep neural network compression and acceleration, and multi-modal learning. 
\end{IEEEbiography}


\begin{IEEEbiography}[{\includegraphics[width=1in,height=1.25in,clip,keepaspectratio]{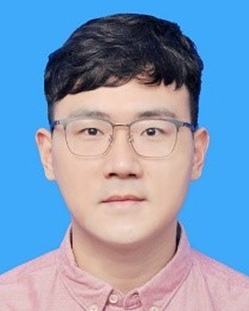}}]{Hang Yu}
(Member, IEEE) received the Ph.D. degree from the University of Technology Sydney, Ultimo, NSW, Australia, in 2020, under the supervision of Prof. Jie Lu.

He is a Full Professor at the School of Computer Engineering and Science, Shanghai University, Shanghai, China. He has authored or co-authored more than 60 publications and his publications have appeared in the \textsc{IEEE Transactions on Knowledge and Data Engineering}, \textsc{IEEE Transactions on Neural Networks and Learning Systems}, \textsc{IEEE Transactions on Cybernetics}, and \textsc{IEEE Transactions ON Fuzzy Systems}. His research interests include streaming data mining, concept drift, and fuzzy systems.

Prof. Yu was awarded the Outstanding Academic Leader of Shanghai. He also regularly serves as a program committee member for numerous national and international conferences.
\end{IEEEbiography}

\end{document}